\documentclass[lettersize,journal]{IEEEtran}
\usepackage{amsmath,amsfonts}
\usepackage{algorithmic}
\usepackage{algorithm}
\usepackage{array}
\usepackage{textcomp}
\usepackage{stfloats}
\usepackage{url}
\usepackage{verbatim}
\usepackage{graphicx}
\usepackage{cite}
\usepackage{times}
\usepackage{epsfig}
\usepackage{graphicx}
\usepackage{amsmath}
\usepackage{amssymb}
 \usepackage{epstopdf}
 \usepackage{color}
\usepackage{xcolor}
\usepackage{booktabs,siunitx} 
\usepackage{makecell}
\usepackage{multirow}
\usepackage{array}
\usepackage{subfigure}
\usepackage{threeparttable}
\usepackage{multirow}
\usepackage{hyperref}
\usepackage{booktabs}
\begin{document}

\title{Detail-recovery Image Deraining via Dual Sample-augmented Contrastive Learning}

\author{Yiyang Shen, Mingqiang Wei, Sen Deng, Wenhan Yang, Yongzhen Wang, Xiao-Ping Zhang,~\IEEEmembership{Fellow,~IEEE,} Meng Wang,~\IEEEmembership{Fellow,~IEEE,} and Jing Qin
 \thanks{Y. Shen, M. Wei and Y. Wang are with the School of Computer Science and Technology, Nanjing University of Aeronautics and Astronautics, Nanjing, China, and also with the MIIT Key Laboratory of Pattern Analysis and Machine Intelligence, Nanjing, China (shenyiyang114@gmail.com; mingqiang.wei@gmail.com; wangyz@nuaa.edu.cn).}
 \thanks{W. Yang is with the School of EEE, Nanyang Technological University, Singapore (wenhan.yang@ntu.edu.sg).}
  \thanks{S. Deng and J. Qin are with The Hong Kong Polytechnic University, Hong Kong SAR, China (sendeng@nuaa.edu.cn; harry.qin@polyu.edu.hk).}
   \thanks{X.-P. Zhang is with the Department of Electrical, Computer and Biomedical Engineering, Ryerson University, Toronto, Canada (xzhang@ee.ryerson.ca).} 
    \thanks{M. Wang is with the School of Computer Science and Information Engineering, Hefei University of Technology, Hefei, China (eric.mengwang@gmail.com).}

}

\markboth{Journal of \LaTeX\ Class Files,~Vol.~14, No.~8, August~2021}%
{Shell \MakeLowercase{\textit{et al.}}: A Sample Article Using IEEEtran.cls for IEEE Journals}


\maketitle

\begin{abstract}

The intricacy of rainy image contents often leads cutting-edge deraining models to image degradation including remnant rain, wrongly-removed details, and distorted appearance. 
Such degradation is further exacerbated when applying the models trained on synthetic data to real-world rainy images. We observe two types of domain gaps between synthetic and real-world rainy images: one exists in rain streak patterns; the other is the pixel-level appearance of rain-free images.
To bridge the two domain gaps, we propose a semi-supervised detail-recovery image deraining network (Semi-DRDNet) with dual sample-augmented contrastive learning. 
Semi-DRDNet consists of three sub-networks:
i) for removing rain streaks without remnants, we present a \textit{squeeze-and-excitation} based rain residual network; 
ii) for encouraging the lost details to return, we construct a \textit{structure detail context aggregation} based detail repair network; to our knowledge, this is the first time; and 
iii) for building efficient contrastive constraints for both rain streaks and clean backgrounds, we exploit a novel \textit{dual sample-augmented contrastive regularization network}.
Semi-DRDNet operates smoothly on both synthetic and real-world rainy data in terms of deraining robustness and detail accuracy.
Comparisons on four datasets including our established Real200 show clear improvements of Semi-DRDNet over fifteen state-of-the-art methods. Code and dataset are available at
\textcolor{magenta}{ \href{https://github.com/syy-whu/DRD-Net}{https://github.com/syy-whu/DRD-Net}}.
\end{abstract}


\section{Introduction}

\begin{figure*}[!t] \centering
	\includegraphics[width=0.8\linewidth]{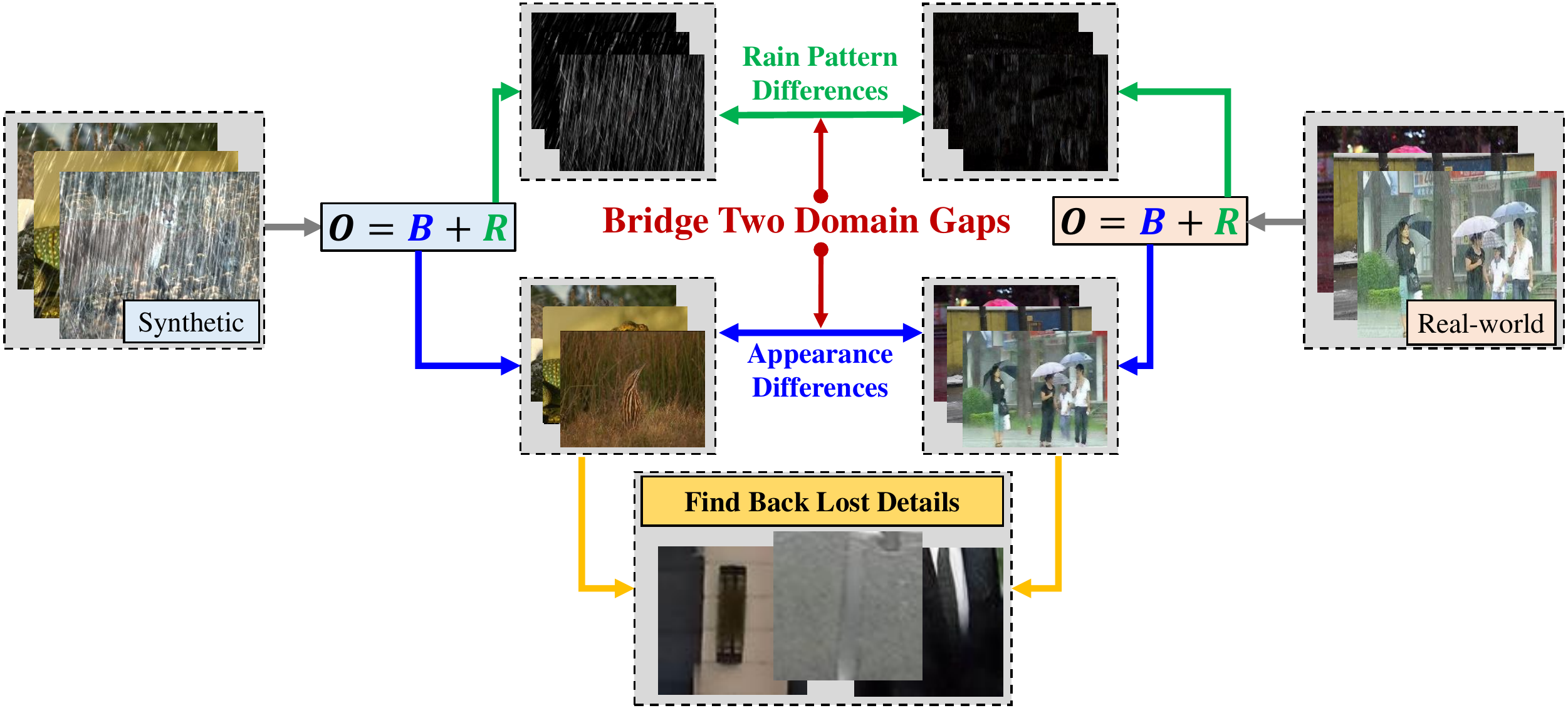}
	\caption{Semi-supervised image deraining is a challenging task. We observe two types of domain gaps between synthetic and real-world rainy images which are not well solved by existing (semi-)supervised image deraining algorithms. Differently, our Semi-DRDNet smoothly bridges the two domain gaps of rain pattern differences and image appearance differences, leading to detail-recovery deraining results in real-world scenarios.}
	
	\label{fig:ab_real}
\end{figure*}

\begin{figure}[!t]
\centering
\subfigure[Real rainy image]{
\includegraphics[width=0.32\linewidth]{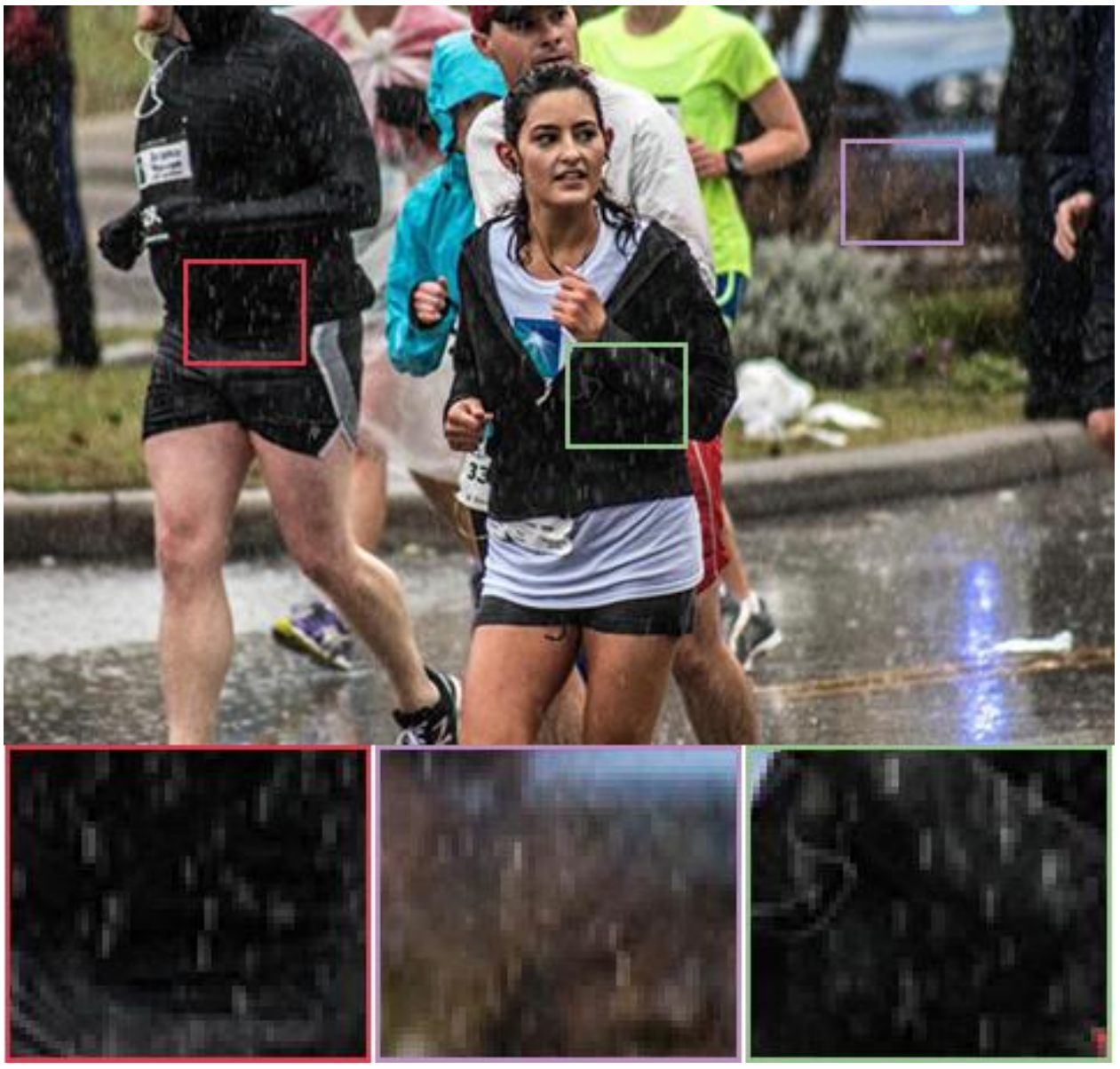}
}
\subfigure[Standard]{
\includegraphics[width=0.32\linewidth]{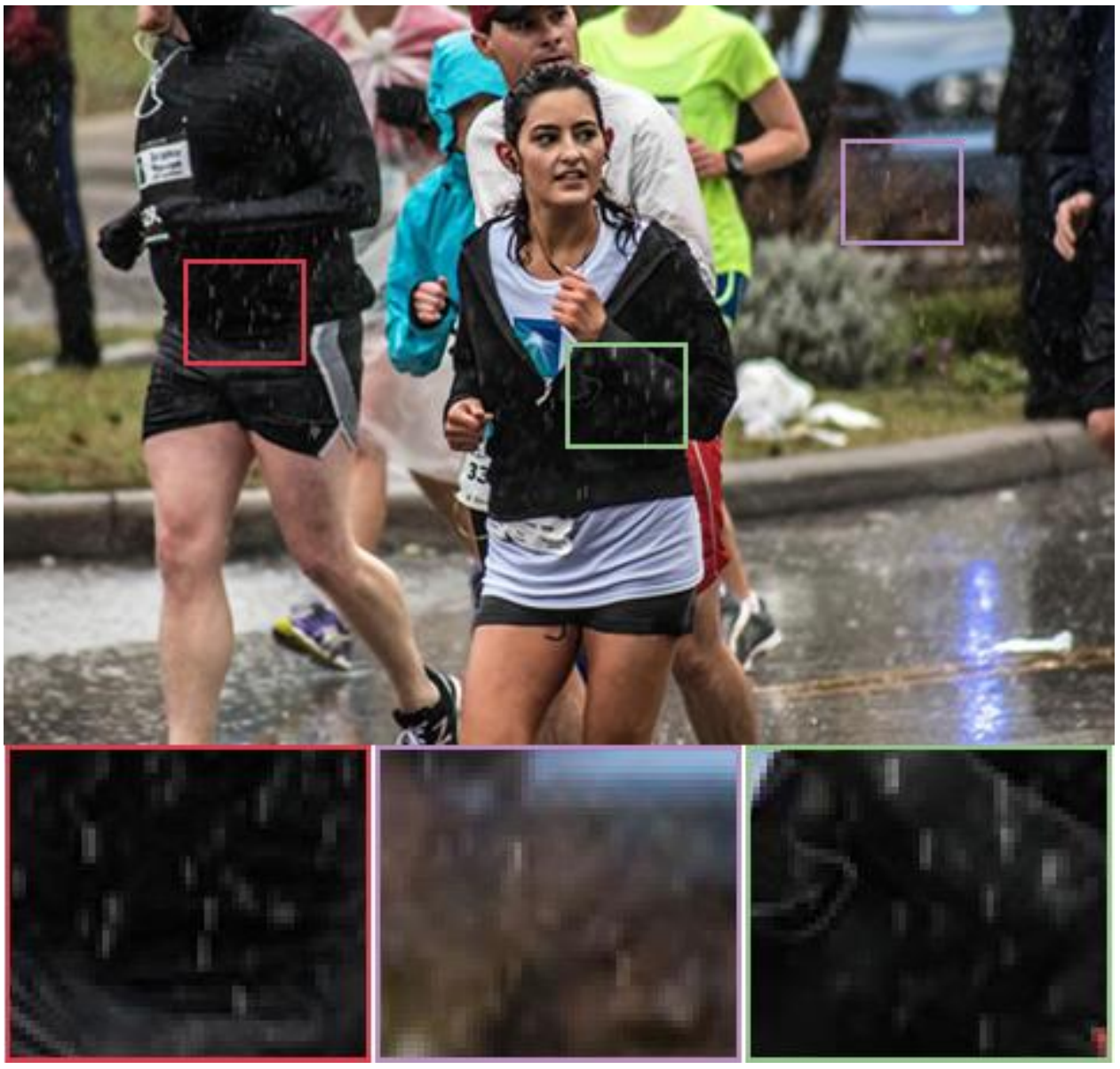}
}
\subfigure[Augmented (ours)]{
\includegraphics[width=0.32\linewidth]{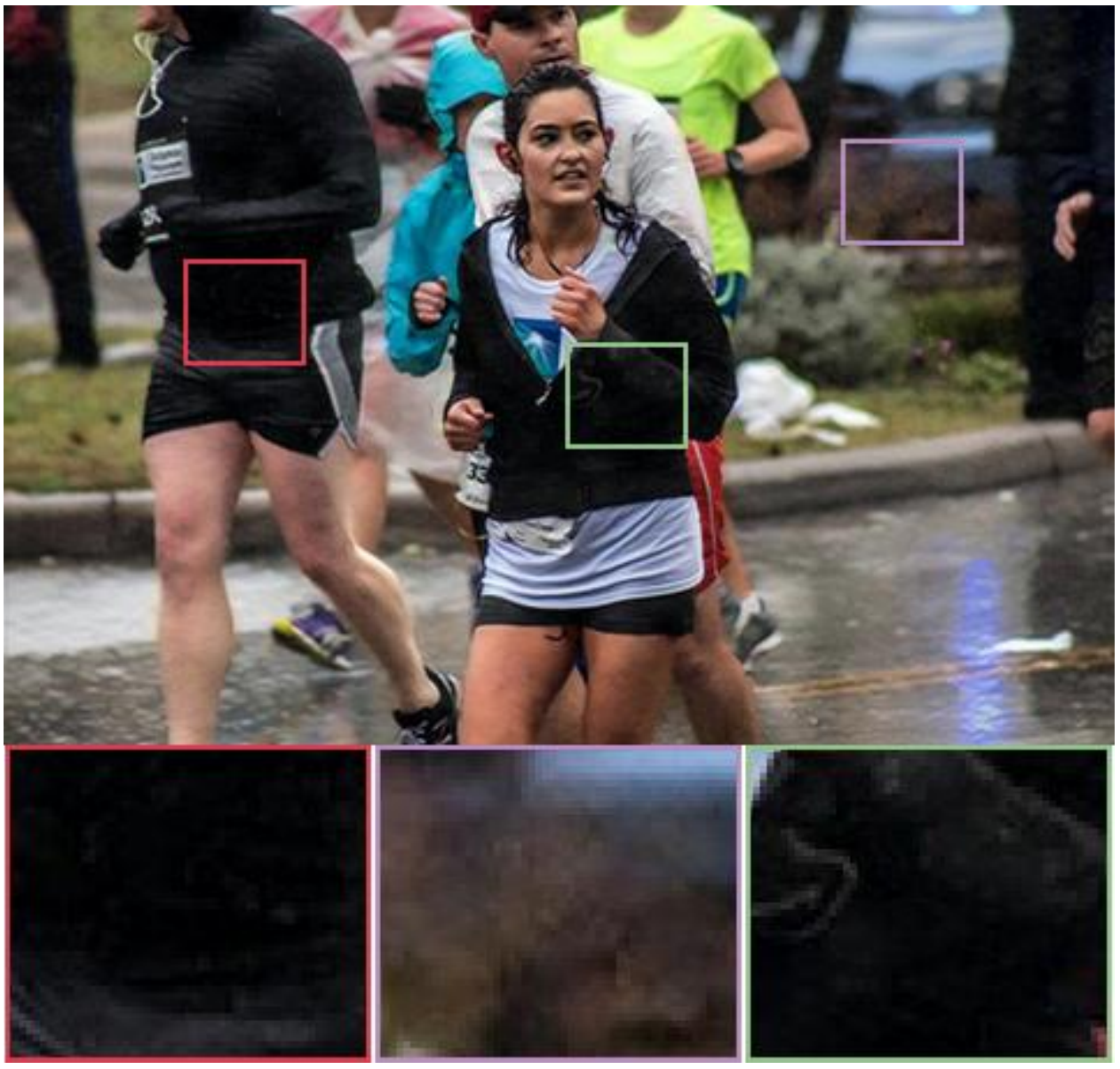}
}
\caption{With dual sample-augmented contrastive learning, Semi-DRDNet generates the detail-recovery deraining result without the remnant of rain streaks (right). In contrast, standard contrastive learning lacks sufficient
considerations of two types of domain gaps during model transfer, leading to a limited generalization ability in real rain
scenarios (middle). }
\label{fig:motivation}
\end{figure}


\par Images captured on rainy days inevitably suffer from the noticeable degradation of visual quality. The degradation causes detrimental impacts on outdoor vision-based systems, such as video surveillance, autonomous driving, and monitoring \cite{liu2022image,hahner2021fog,bijelic2020seeing}. It is, therefore, indispensable to remove rain in rainy images, which is referred to as \textit{image deraining}.


Image deraining is an ill-posed problem. Traditional methods of image deraining usually exploit various image priors, such as Gaussian mixture model \cite{Gaussian_mixture_model}, sparse coding \cite{sparse_coding, sparse_code} and low-rank representation \cite{low_rank, guo2019robust}. For this wisdom of image deraining, users have to tweak parameters multiple times to obtain satisfactory results. Such inconvenience discounts the efficiency and user experience in medium- and heavy-rain removal in practical scenarios. 
Recent efforts tend to construct various rainy datasets \cite{yang2017deep,zhang2019image,zhang2018density,li2019heavy,shen2022mba} and design novel neural networks \cite{detail_layer,wang2021multi,yang2021recurrent,yang2019joint,quan2021removing,wang2020rethinking} for automatic yet quality image deraining. However, these datasets are synthetic and paired, which insufficiently reflect the real-world rain that varies largely in scales (e.g., heavy/moderate/light) and directions. Therefore, leveraging cutting-edge deraining models trained on these datasets unquestionably leads to poor performance on real-world rainy images.

Semi-supervised deraining techniques, as promising solutions, 
leverage paired synthetic data for good initialization and unpaired real-world data for generalization \cite{huang2021memory,ye2021closing,yasarla2020syn2real,wei2019semi,wei2021semi,yasarla2021semi}. 
However, image degradation (e.g., the loss of image details, remnant rain, halo artifacts, and/or color distortion) is common to observe in these semi-supervised methods. We analyze why the predicament of image degradation still happens during deraining.
Naturally, the ultimate goal of image deraining is to recover the clean background $\textbf{B}$ from its observation $\mathbf{O = B + R}$ with $\textbf{R}$ as the rain layer:
\begin{itemize}
\item $\textbf{B}$ contains image details similar to $\textbf{R}$ in scale. That is, the magnitude of image details is similar to and even smaller than that of rain streaks. No state-of-the-art semi-supervised methods can serve as a real-world deraining panacea for various applications: they produce results with a trade-off between rain removal and detail preservation.
\item $\textbf{B}$ and $\textbf{R}$ are both unknown. That is, two types of domain gaps exist between synthetic and real-world rainy images. The first type of domain gap falls into $\textbf{R}$ where synthetic rain patterns cannot cover a comprehensive range of real-world rain patterns. The second falls into $\textbf{B}$ where the pixel-level appearance differences exist between synthetic and real-world clean backgrounds, including but not limited to illumination, color bias, and camera noise.
\end{itemize}

To struggle with image degradation, we propose a semi-supervised detail-recovery image deraining network, namely Semi-DRDNet, based on dual
sample-augmented contrastive learning (see Fig. \ref{fig:ab_real}).
Semi-DRDNet consists of three sub-networks: (i) A squeeze-and-excitation (SE)-based rain residual network (RRN) to separate rain streaks from input images. SE aggregates feature maps in the same convolutional layer to make full advantage of spatial contextual information for complete rain removal. 
(ii) A detail repair network (DRN) to encourage the lost details to return to the image after deraining by RRN. Especially, we present the structure detail context aggregation block (SDCAB), which has larger reception fields and makes full use of the rain-free image patches to facilitate detail recovery. 
(iii) A dual sample-augmented contrastive regularization network (DualSCRNet) to bridge two types of domain gaps between synthetic and real-world rainy images. To achieve this, unlike standard contrastive regularization networks \cite{wu2021contrastive,park2020contrastive,ye2022unsupervised,chen2021unpaired}, we propose two core sample-augmented strategies to build efficient contrastive constraints for both rain streaks and clean backgrounds.
First, to overcome the domain gap between synthetic and real-world rain layers, we augment negative samples (rainy images) to cover a sufficiently comprehensive range of rain streak patterns. 
Second, to overcome the domain gap between synthetic and real-world clean background layers, we augment positive samples (clean images) by mapping unpaired clean images from their original domain distribution to a new domain distribution related to real-world data. Benefiting from dual sample-augmented contrastive learning, Semi-DRDNet can produce more promising results in real-world scenarios (see Fig. \ref{fig:motivation}).

Experiments on four benchmark datasets show that Semi-DRDNet outperforms fifteen methods both quantitatively and qualitatively.
The main contributions are as follows:
\begin{itemize}
	\item We propose Semi-DRDNet, a semi-supervised detail-recovery image deraining network via dual sample-augmented contrastive learning. Unlike existing methods, Semi-DRDNet smoothly bridges two types of domain gaps between synthetic and real-world rainy images. 
	\item Semi-DRDNet consists of three sub-networks. The first two sub-networks are parallel and trained on paired synthetic data. The third branch is connected with the first two sub-networks in a cascaded way for semi-supervised training.
	\item We propose a squeeze-and-excitation (SE)-based rain residual network (RRN) to exploit spatial contextual information for complete rain removal. 
	\item We propose a detail repair network (DRN) to encourage the lost details back to the image after deraining by RRN. \textit{To our knowledge, this is the first time}.
	\item We observe two domain gaps in real-world image deraining tasks, and propose a dual sample-augmented contrastive regularization network, termed a DualSCRNet, to improve the generalization ability of Semi-DRDNet. \textit{To our knowledge, this is the first time}.
\end{itemize}
\par \textbf{Difference from our conference paper:} This work covers and extends our conference version DRD-Net \cite{deng2020detail} from the following aspects: 
(i) We generalize DRD-Net to Semi-DRDNet, a new semi-supervised image deraining paradigm to bridge two types of domain gaps between synthetic and real-world rainy data.
Thus, our previous DRD-Net can be seen as a simplified version of Semi-DRDNet. 
(ii) We propose a dual sample-augmented contrastive regularization network, which leverages augmented positive samples (clean images) and negative samples (rainy images) to build a more generalized and discriminative semi-supervised deraining paradigm.
(iii) Both the proposed detail recovery network and dual sample-augmented contrastive regularization network can be detachable and incorporated into existing deraining methods, e.g., \cite{yasarla2020syn2real,ye2021closing},
to boost their performance. 
(iv) Unlike the previous DDN-SIRR dataset \cite{wei2019semi} which only contains 147 real-world rainy images, we establish a new real-world rainy dataset, called Real200. Real200 contains 2000 real-world rainy images (1800 images for training and 200 images for testing) from a wide range of real-world scenes. Thus, we conduct more experiments on the synthetic and real-world datasets to verify the superior performance as compared to existing methods.
(v) Our results show clear improvements over its previous version, i.e., DRD-Net \cite{deng2020detail} on both synthetic and real-world rainy images.

\section{Motivations}
Image degradation, such as remnant rain, wrongly-removed details, and distorted appearance, happens when applying cutting-edge deraining models on rainy images, due to the intricacy of rainy image contents. Such degradation is further exacerbated if applying the models trained on synthetic data to real-world rainy images.
At the top level, it is natural to i) train two parallel networks for rain removal and image detail recovery; and ii) train an additional network for bridging the domain gap between synthetic and real-world rainy images.
Such a learning paradigm can leverage both synthetic and real-world rainy images to obtain better deraining results. Especially, given any synthetic/real-world rainy image as input, the network is expected to output a deraining result without both remnant rain and the loss of image details.

\begin{figure*}[!t] \centering
	\includegraphics[width=1\linewidth]{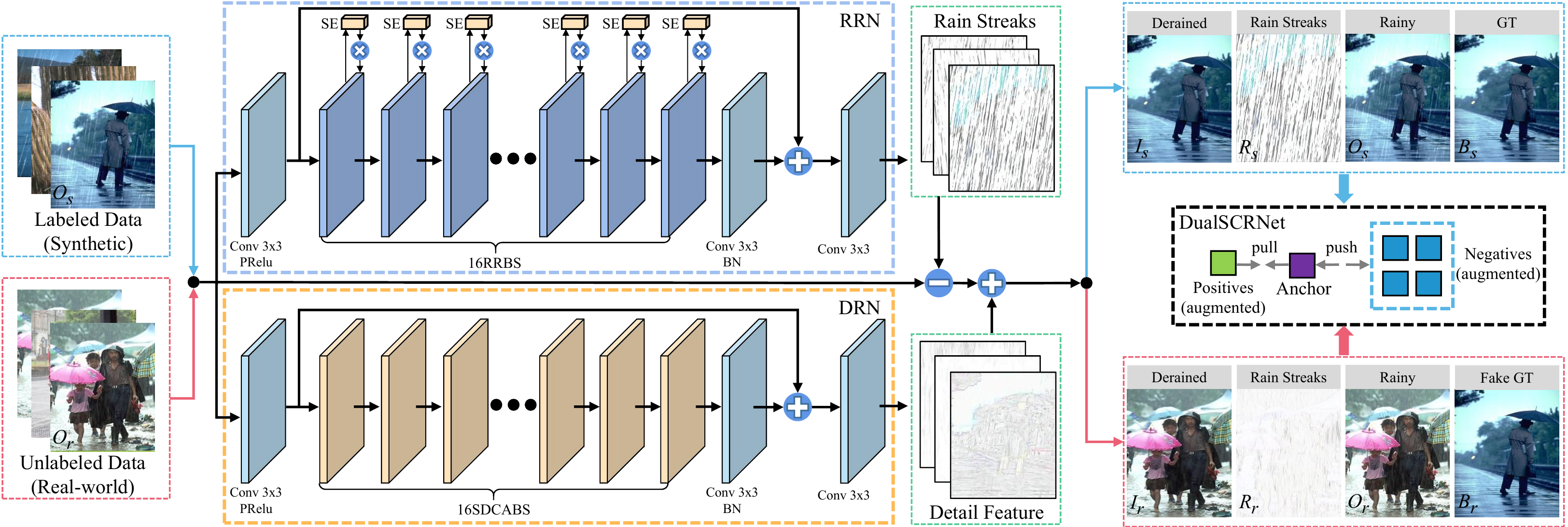}
	\caption{Pipeline of our Semi-DRDNet. Semi-DRDNet consists of three sub-networks, i.e., the rain removal network (RRN), the detail repair network (DRN), and the dual sample-augmented contrastive regularization network (DualSCRNet).
	RRN and DRN are parallel and then connected to DualSCRNet in a cascaded way. Semi-DRDNet can remove rain and recover image details, and then be transferred to handle real-world data without labels. In detail, RRN, which combines the squeeze-and-excitation (SE) operation with residual blocks to make full advantage of spatial contextual information, aims at removing rain streaks from the rainy images. DRN, which integrates the structure detail context aggregation block (SDCAB) to aggregate context feature information from a large reception field, seeks to recover the lost details to the derained images. 
	DualSCRNet, which utilizes contrastive learning to obtain the information of augmented rainy and clean images as negative and positive samples, encourages the derained images and the clean images to pull together in the embedding space while pushing them away from the rainy images.
	Note that $O_{s}$ represents the paired synthetic rainy image with the ground truth $B_{s}$. $I_{s}$ and $R_{s}$ represent the derained image and the synthetic rain streaks layer, respectively. Similarly, $O_{r}$ represents the unpaired real-world rainy image with the `fake' ground truth $B_{r}$, which is randomly selected from clean images of synthetic data. $I_{r}$ and $R_{r}$ represent the derained image and the real-world rain streaks layer, respectively.}
	\label{fig:network framework}
\end{figure*}


\textbf{Motivation 1.} The current wisdom of image deraining usually leads to the loss of image details, since rain streaks and image details are both of high frequency in nature and they inevitably share similar geometrical properties.
We find that most of the existing image deraining networks pay little attention to recovering image details, once they are lost during deraining.

An effective image deraining network should involve the estimation of
two components: rain streaks and image details. However, learning both components simultaneously by a single network is somewhat challenging. This motivates us to leverage an additional network, i.e., the detail repair network (DRN),
to facilitate the single deraining networks for image detail recovery.

\textbf{Motivation 2.} 
The aforementioned supervised two sub-networks (one for rain removal and the other for detail recovery) trained on synthetic rainy datasets still operate poorly on real-world rainy images. 
Semi-supervised image deraining is a new paradigm that leverages paired synthetic data for model initialization and unpaired real-world data for model generalization. For example, if denoting the rain-free images as positive samples, and the rainy images as negative samples, one can potentially learn to pull the derained images together with the positives and push them apart from the negatives in a representation space \cite{wu2018unsupervised,he2020momentum,chen2020simple,saunshi2019theoretical,wu2021contrastive}. However, current wisdom of such paradigm lacks sufficient considerations of two types of domain gaps during model transfer, leading to a limited generalization ability in real-world rain scenarios (see Fig. \ref{fig:motivation}).
To build efficient contrastive constraints, we consider two novel sample-augmented strategies to build augmented contrastive constraints for a better performance of real-world image deraining. 
First, we augment negative samples (rainy images) to cover a sufficiently comprehensive range of rain streak patterns. Second, we augment positive samples
(clean images) to generate novel domain distributions for unpaired clean images. With dual sample-augmented strategies, Semi-DRDNet can generate a more pleasing result (see Fig. \ref{fig:motivation}).

\section{Semi-DRDNet}
We propose a semi-supervised detail-recovery image deraining network (Semi-DRDNet). The pipeline of Semi-DRDNet is depicted in Fig. \ref{fig:network framework}. For both rain removal and detail recovery of single images, different from existing solutions, Semi-DRDNet consists of three sub-networks. 
First, we introduce a rain residual network (RRN) to train a function that maps the rainy images to their rain streaks. Therefore, we can obtain the preliminary derained images by separating the rain streaks from the rainy images. Second, different from other methods which try to decompose a single rainy image into a background layer and a rain streak layer, we present an additional detail repair network (DRN) to get back the lost details. Third, we present a novel dual sample-augmented contrastive regularization network (DualSCRNet) to improve the quality of real-world derained images.


\begin{figure*}[!t] \centering
	\includegraphics[width=0.9\linewidth]{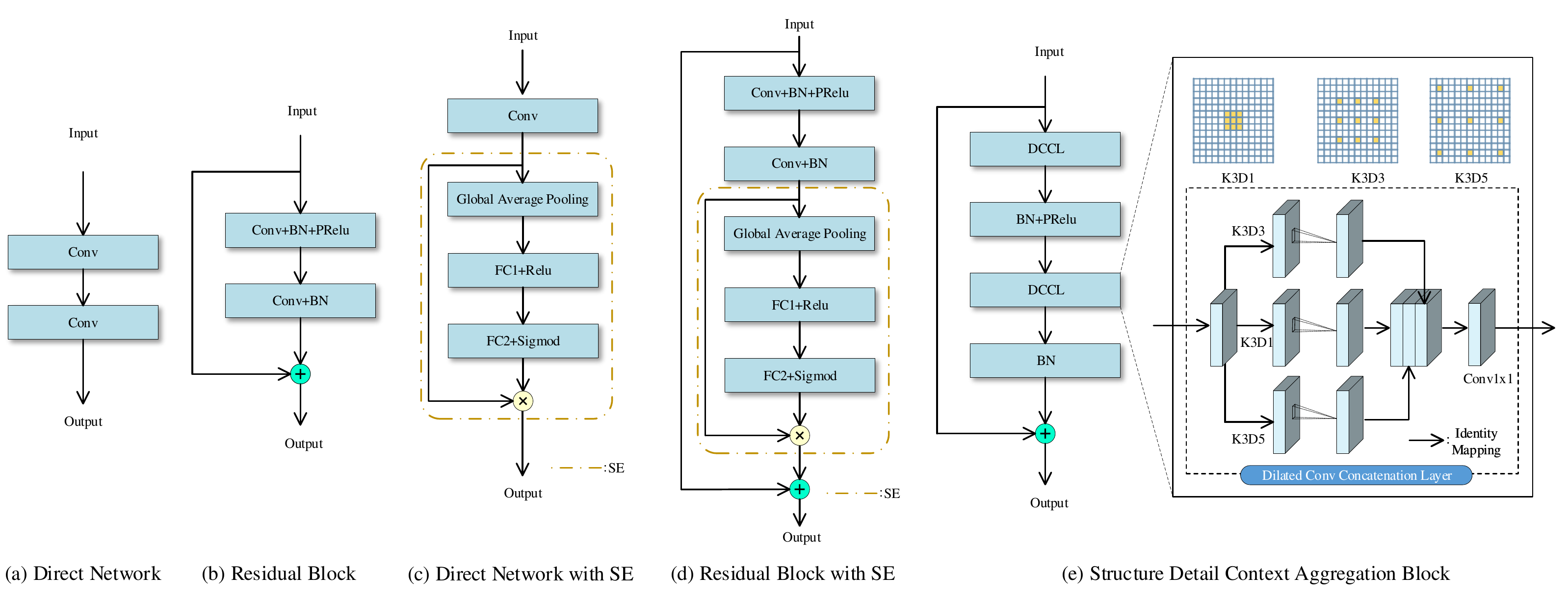}
	\caption{Different convolution styles. From (a)-(d): (a) direct network, (b) residual block, (c) direct network with SE \cite{rescan}, (d) rain residual block with SE used in our rain residual network, and (e) structure detail context aggregation block used in our detail repair network.}
	\label{fig:different connect}
\end{figure*}
\subsection{Rain Residual Network}

\par Residual learning is a powerful tool for image restoration tasks like denoising and deraining \cite{dncnn, rescan}. Since rain streaks are sparser than the rain-free background scene \cite{rescan}, we develop a rain residual network (RRN) to map the rainy image to rain streaks. 
\par Our RRN utilizes Squeeze-and-Excitation (SE) \cite{SE} (see the top part of Fig. \ref{fig:network framework}). Considering that the skip-connections can provide long-range information compensation and enable the residual learning \cite{li2018non}, we combine SE with the residual block in our RRN, which is different from Fig. \ref{fig:different connect}(c) used in RESCAN \cite{rescan}. RRN includes 3 convolution layers and 16 rain residual blocks. The first layer can be interpreted as an encoder, which is used to transform the rainy image into the feature maps, and the last two layers are used to recover the RGB channels from feature maps.

\par Mathematically, the rain residual block is formulated as
\begin{equation}
RRB = SE(Res(\mathbf{X_{0}})),
\end{equation}
where $RRB$ is the output of the rain residual block, $SE(\cdot)$ and $Res(\cdot)$ denote the SE operation and the residual block as shown in Fig. \ref{fig:different connect}(d) respectively, and $\mathbf{X_{0}}$ is the input signal.

\begin{figure}[!t] \centering
	\includegraphics[width=0.9\linewidth]{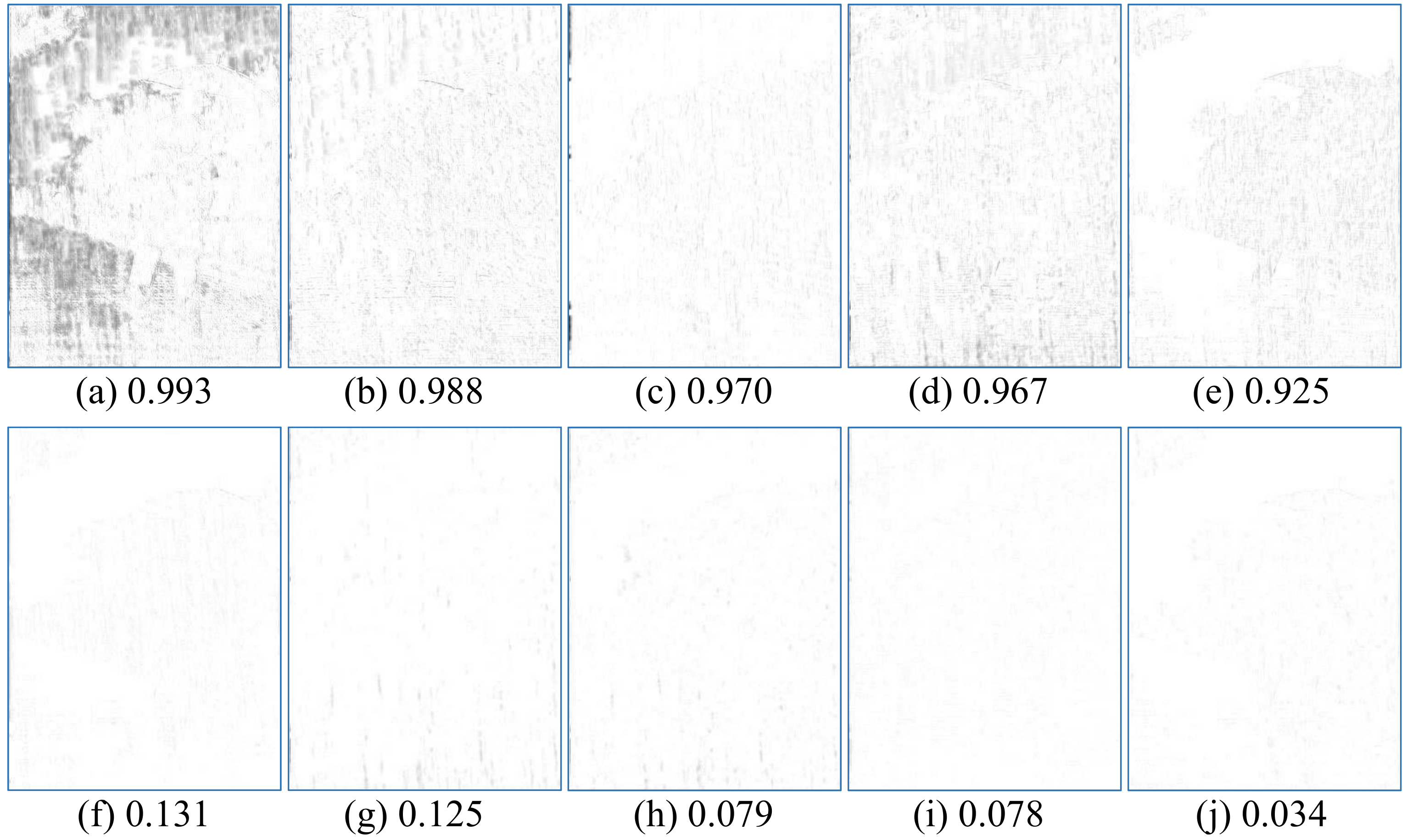}
	\caption{Feature maps with different weights. The images in (a)-(e) denote the top five high weighted feature maps, and in (f)-(j) denote the top five low weighted feature maps. 
	}
	\label{fig:se}
\end{figure}

Spatial contextual information is effective in image deraining \cite{huang2012context, rescan}. Nevertheless, the different feature channels in the same layer are independent and have few correlations during the previous convolution operation. A main difference from the common residual block is that we combine SE into the residual block. Since SE models a correlation between different feature channels, we can intensify the feature channel to have more context information by giving a larger weight. Conversely, the feature channels with less spatial contextual information will just receive a small weight. All the weights of different channels are learned by RRN automatically during the training steps. To obtain insight into the correlation between the SE weight and the content of layers, we visualize the feature maps with different weights as shown in Fig. \ref{fig:se}. It is clear that the feature maps with more spatial contextual information have received a higher weight as expected.

\begin{figure*}[!t] \centering
	\includegraphics[width=0.9\linewidth]{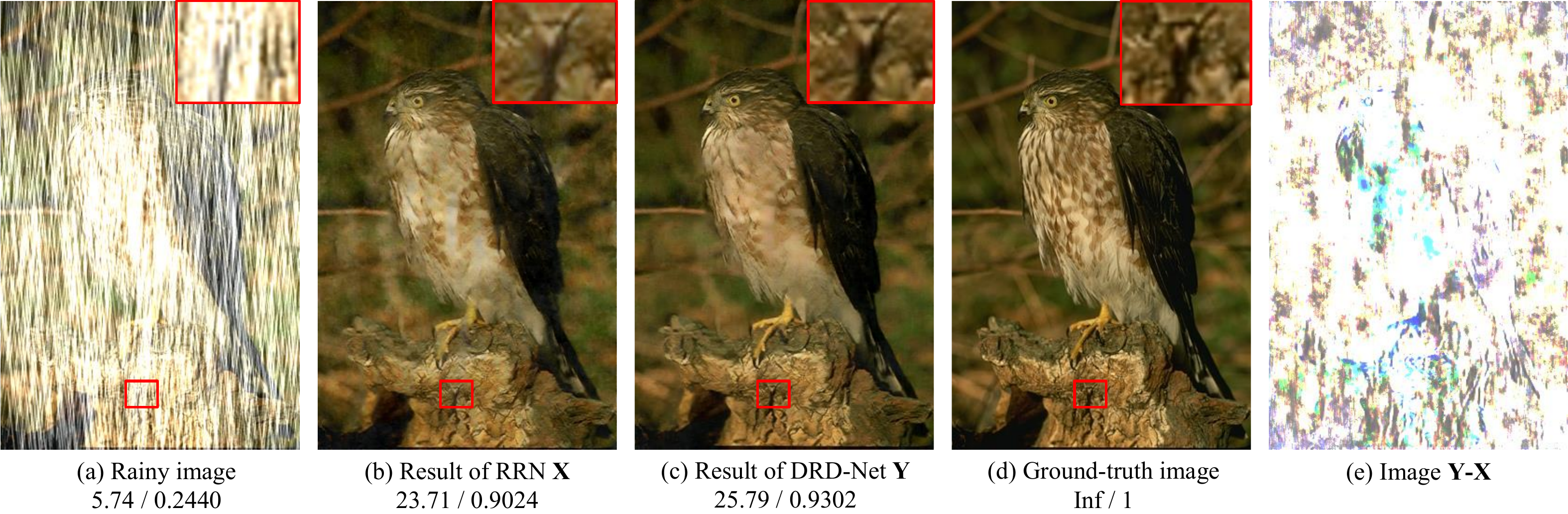}
	\caption{Validation of our DRN for image detail recovery during deraining. From (a)-(e): (a) the input rainy image, (b) the result \textbf{X} by only using our RRN, (c) the result \textbf{Y} by DRD-Net (RRN+DRN), (d) GT, and (e) the image of \textbf{Y-X}. 
	}
	\label{fig:Y-X}
\end{figure*}

\subsection{Detail Repair Network}

\par Image deraining leads to image degradation in nature. We can train an additional detail-recovery network (DRN) that makes the detail-lost images reversible to their artifact-free status. Inspired by \cite{Yang_dl}, we design our DRN based on the structure detail context aggregation block (SDCAB). The difference from \cite{Yang_dl} is that we adopt SDCAB into the whole network flow to utilize multi-scale features, while \cite{Yang_dl} only applies the multi-scale dilated blocks in the first layer to extract image features. This modification benefits our DRN well. Specifically, SDCAB consists of different scales of dilated convolutions and $1 \times 1$ convolutions as shown in Fig. \ref{fig:different connect}(d). Since a large receptive field is very helpful to acquire much contextual information \cite{rescan}, we present 3 dilated convolutions whose dilation scales are 1, 3 and 5 in SDCAB, respectively. Then, in order to extract the most important features, we concatenate the output of dilated convolutions and utilize a $1 \times 1$ convolution to reduce the feature dimensions. For reducing the complexity in training, the residual network is also introduced into SDCAB.

\par As shown in Fig. \ref{fig:different connect}(d), the dilated convolution concatenation layer (DCCL) can be expressed as
\begin{equation}
\begin{split}
DCCL = Conv_{1 \times 1}(Cat[Conv_{3\times 3, d_{1}}(X), \\
Conv_{3\times 3, d_{3}}(X), Conv_{3\times 3, d_{5}}(X)]),
\end{split}
\end{equation}
where $Conv_{x \times x, d_{y}}$ denotes the dilated convolutions with the kernel size of $x \times x$, and the dilation scale is $y$. $Cat(\cdot)$ is a concatenation operation and $X$ is the input feature.
\par Mathematically, SDCAB is formulated as
\begin{equation}
SDCAB = Add[X_{input}, BN(DCCL_{2})],
\end{equation}
where 
\begin{equation}
DCCL_{2} = PRelu(BN(DCCL_{1}(X_{input}))).
\end{equation}



A large receptive field plays an important role in obtaining more information. With a larger receptive field, we obtain more context information to find back the lost details. 
One knows from Fig. \ref{fig:Y-X} that DRN finds back the details that were lost by filtering the rainy image to obtain the final derained image $X$. 
We provide more experimental results on three datasets to compare the performance of image deraining with/without DRN in Table \ref{Tab:ablation_1}: our Semi-DRDNet outperforms other network architectures, thanks to the DRN capability to ﬁnd back the lost details.

\textbf{Relationship between our SDCAB and MSARR in \cite{Yang_dl}.} 
The similarity between our SDCAB and multi-scale aggregated recurrent ResNet (MSARR) in \cite{Yang_dl} is the use of dilated convolution, while the differences lie in several aspects: (i) In \cite{Yang_dl}, the dilated convolution is applied only once to extract features from the original image. In contrast, SDCAB is composed of several dilated convolution concatenation layers (DCCLs), and our network employs several SDCABs. Such a structure
further enlarges the receptive field to
capture non-local correlations among details. (ii) In \cite{Yang_dl}, the features extracted by parallel dilated convolution layers are simply added together, while those extracted by DCCL are concatenated and combined with automatically adjusted weights. This benefits the utilization of features from different dilated scales and the localization of image
details. (iii) The skip connection is combined with DCCL to construct SDCAB, which not only helps reuse the previous features and explore new ones but also prevents this deep structure from gradient vanishing.

\begin{figure*}[!t] \centering
	\includegraphics[width=1\linewidth]{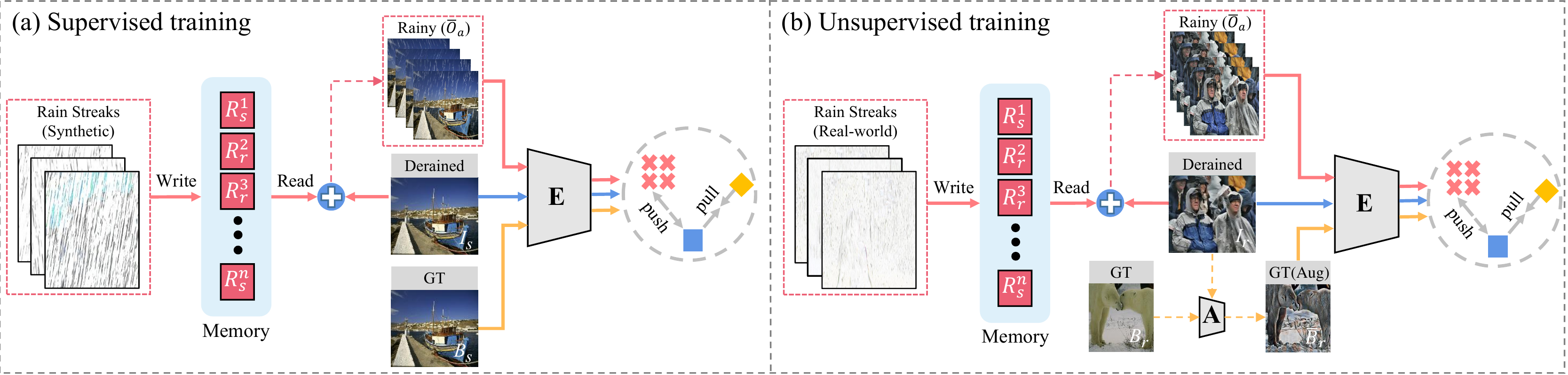}
	\caption{Structure of DualSCRNet. It consists of the supervised training phase and the unsupervised training phase.}
	\label{fig:SCRN}
\end{figure*}
\subsection{Dual Sample-augmented Contrastive Regularization Network}
To bridge two types of domain gaps between synthetic and real-world rainy images, we leverage a dual sample-augmented contrastive regularization network (DualSCRNet). 
The goal of contrastive learning is to learn a representation to pull ``positive" pairs in the embedding space and push apart the representation between ``negative" pairs.
To incorporate contrastive learning into our semi-supervised model for improving the performance and the generalization capability in real-world applications, we consider the following aspects: to build the pairs of ``positive" and ``negative" samples, to find the suitable latent feature space of pairs to train the network, and to build efficient contrastive constraints for both rain streaks and clean
backgrounds.

\textbf{Supervised training phase:} In this phase, we use paired synthetic data for training. We choose the synthetic derained image $I_{s}$ as the anchor (see Fig. \ref{fig:SCRN}). 
The synthetic rainy image $O_{s}$ has a corresponding clean image $B_{s}$, which is selected as the positive sample. 
There is a simple fact that real-world rains include a broader range of rain streak patterns than synthetic ones. To overcome the domain gap, we develop a new sample-augmented strategy by augmenting negative samples (rainy images) to cover complex real-world rain streak patterns.
Concretely, we employ a special queue-style memory bank to record various types of rain streak layers generated by RRN in both un-/supervised training phases. 
Then, we randomly read the rain streak layers from the memory bank to generate rainy images with different patterns as
\begin{equation}
O_{a}^{k} = I_{s}+R^{k},
\end{equation}
where $O_{a}^{k}$ denotes the $k$-th generated rainy image, and $R^{k}$ represents the $k$-th synthetic or real-world rain streak layer from the memory bank. 
Finally, we use the original rainy image $O_{s}$ and the generated rainy image $O_{a}$ to build an augmented rainy images set $\bar{O}_{a}=\{O_{a}^{1},O_{a}^{2},O_{a}^{3},...,O_{a}^{m-1},O_{s}\}$ as negative samples to cover various types of rain streak patterns.
\begin{figure}[!t] \centering
	\includegraphics[width=1\linewidth]{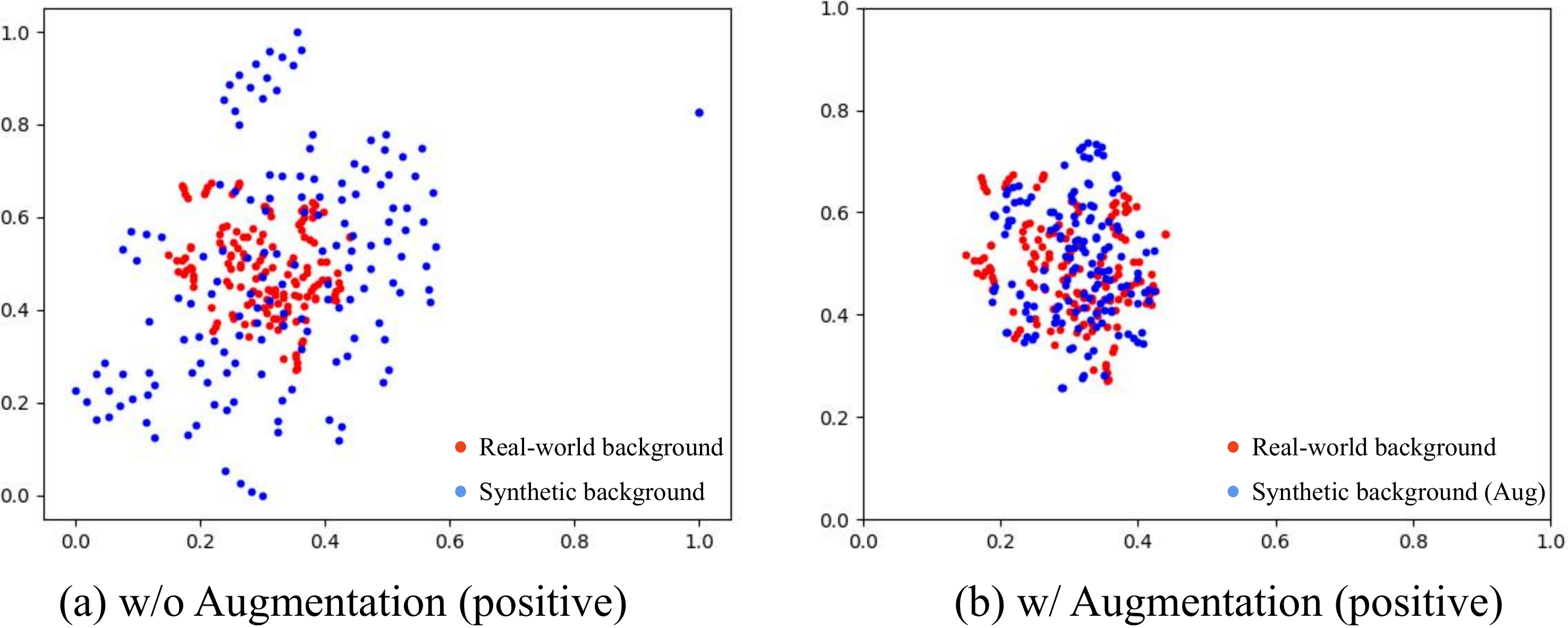}
	\caption{t-SNE visualization of extracted real-world clean background  features (red) and synthetic clean background features (blue). 
Note that the embeddings of the real-world background features and the augmented synthetic background features have lower distances in the latent space than without augmentation.}
	\label{fig:tsne}
\end{figure}
We select a pre-trained VGG-16 to extract the common intermediate feature for the latent space. The supervised contrastive loss is defined as
\begin{equation}
L_{Dual(sup)} = \sum_{k=1}^{m}\sum_{i=1}^{n}\omega_{i} \bullet \frac{\mathop{ {\parallel} } \varphi_{i}(B_{s})- \varphi_{i}(I_{s}) \mathop{ {\parallel} }_{2}^{2}}{\mathop{ {\parallel} } \varphi_{i}(\bar{O}^{k}_{a})- \varphi_{i}(I_{s}) \mathop{ {\parallel} }_{2}^{2}},
\end{equation}
where $\varphi_{i}(\cdot)$, $i = 1,2,...,n$, refers to extracting the $i$-th hidden features from the pre-trained VGG-16 network. Herein we choose the 2-nd, 3-rd, and 5-th max-pooling layers. $\omega_{i}$ are weight coefficients, and we set $\omega_{1}=0.2$, $\omega_{2}=0.5$, and $\omega_{3}=1$. $\bar{O}^{k}_{a}$, $k= 1,2,3,4$, refers to the $k$-th augmented rainy image.




\textbf{Unsupervised training phase:} 
Existing approaches \cite{wu2021contrastive,park2020contrastive} usually train a contrastive network on the paired positive and negative samples. 
However, pairs of real-world rainy images and clean images are intractable to obtain. Thus, we explore how to use unpaired real-world data for training.
We first choose the real-world derained image $I_{r}$ as the anchor (see Fig. \ref{fig:SCRN}). Different from the supervised phase, we only read real-world rain streak layers from the memory bank to generate an augmented rainy images set $\bar{O}_{a}=\{O_{a}^{1},O_{a}^{2},O_{a}^{3},...,O_{a}^{m-1},O_{r}\}$ as negative samples.

Unlike paired synthetic data, real-world rainy images have no corresponding ground truths. It is reasonable to select unpaired clean background images generated from paired synthetic datasets as ``fake or pseudo labels" for the unsupervised training phase \cite{wei2021semi,ye2021closing,liu2021unpaired,dong_liang}.
However, we observe that there is a domain gap between synthetic and real-world clean backgrounds.
To overcome this domain gap, we design a new sample-augmented strategy by transferring these unpaired clean images to a domain distribution related to real-world data.
Concretely, given the real-world derained image $I_{r}$ and the unpaired clean image $B_{r}$, we adopt a domain transformation \cite{liu2021adaattn} to map $B_{r}$ from its source domain to the target domain of $I_{r}$ for positive sample augmentation as
\begin{equation}
\bar{B}_{r} = A(F_{c}(B_{r}),F_{s}(I_{r})),
\end{equation}
where $A(\cdot)$ denotes a domain transformation, which is usually seen as a way of simulating domain shift. 
$F_{c}(\cdot)$ and $F_{s}(\cdot)$ denote encoders to obtain content and style features, respectively. To better understand the effect of such an augmented strategy, we
further use t-SNE \cite{van2008visualizing} to visualize features in Fig. \ref{fig:tsne}. We observe that the real-world and augmented synthetic clean background layers completely overlap in the latent space, which denotes that our positive augmented strategy can bridge the domain gap between synthetic and real-world clean backgrounds.
Thus, we choose the augmented clean image $\bar{B}_{r}$ as the positive sample, which helps DualSCRNet to generate more discriminative visual mappings.
The unsupervised contrastive loss is formulated as
\begin{equation}
L_{Dual(unsup)} = \sum_{k=1}^{m}\sum_{i=1}^{n}\omega_{i} \bullet \frac{\mathop{ {\parallel} } \varphi_{i}(\bar B_{r})- \varphi_{i}(I_{r}) \mathop{ {\parallel} }_{2}^{2}}{\mathop{ {\parallel} } \varphi_{i}(\bar{O}^{k}_{a})- \varphi_{i}(I_{r}) \mathop{ {\parallel} }_{2}^{2}},
\end{equation}
where we keep the same setting as the supervised phase.

\subsection{Comprehensive Loss Function}
The comprehensive loss function is formulated as
\begin{equation}
L_{total} = L_{sup}+\lambda_{unsup}L_{unsup},
\end{equation}
where $\lambda_{unsup}$ is a pre-defined weight to control the contribution from $L_{sup}$ and $L_{unsup}$.

\textbf{Supervised training phase:} In this phase, we use the labeled synthetic data to learn the network parameters. Specifically, we minimize the supervised loss function as
\begin{equation}
L_{sup} = L_{d}+\lambda_{r}L_{r}+\lambda_{dual}L_{Dual(sup)},
\end{equation}
where $L_{r}$, $L_{d}$ and $L_{Dual(sup)}$ are the rain residual loss, the detail repair loss, and the supervised contrastive loss, respectively. 
$L_{r}$ and $L_{d}$ are defined as
\begin{equation}
L_{r} = \mathop{ {\parallel} } f(\mathbf{O}_{s})-\hat{\textbf{R}}\mathop{ {\parallel_{1}} },
\end{equation}
\begin{equation}
L_{d} =\mathop{ {\parallel} }(\hat{\mathbf{I}}_{s} + g(\mathbf{O}_{s})-\mathbf{B}_{s}\mathop{ {\parallel_{1}} },
\end{equation}
where $f(\cdot)$ and $g(\cdot)$ are the functions of RRN and DRN respectively, $\mathbf{O}_{s}$ is the synthetic rainy image, $\mathbf{\hat{\textbf{R}}}$ is the ground-truth rain streak layer obtained by subtracting the ground truth $\mathbf{B}_{s}$ from the rainy image $\mathbf{O}_{s}$, and $\hat{\mathbf{I}}_{s}$ is the preliminary derained image obtained by subtracting the generated rain streaks $\textbf{R}_{s}$ from $\mathbf{O}_{s}$.

\textbf{Unsupervised training phase:} We leverage the unlabeled real-world data to improve the generalization performance. Specifically, we minimize the unsupervised loss function as
\begin{equation}
L_{unsup} = L_{Dual(unsup)},
\end{equation}
where $L_{Dual(unsup)}$ is the unsupervised contrastive loss. Compared to complex unsupervised loss functions of existing semi-supervised deraining models, we only adopt the unsupervised contrastive loss in this phase and can achieve satisfactory results.
By bridging the two domain gaps via DualSCRNet, our Semi-DRDNet handles the challenging real-world rainy case effectively. It not only safeguards the result from rain remnants, but also protects weak image details, which is different from its (semi-)supervised competitors including DRD-Net.

\section{Experiment and Discussions}

\begin{table*}[!t]
	\centering
	\caption{Quantitative experiments evaluated in Rain200L, Rain200H and Rain800. All the supervised methods are directly trained on Rain200L, Rain200H and Rain800, and the semi-supervised ones are trained on Rain200L$\&$Real200, Rain200H$\&$Real200 and Rain800$\&$Real200.}
	\label{Tab:compare}
	\begin{threeparttable}
		\small
		\centering
		\setlength{\tabcolsep}{7mm}{
			\begin{tabular}{ccccccc}
				\toprule
				\multirow{2}{*}{Dataset}&
				\multicolumn{2}{c}{Rain200L}&\multicolumn{2}{c}{Rain200H}&\multicolumn{2}{c}{Rain800}\cr
				\cmidrule(lr){2-3} \cmidrule(lr){4-5} \cmidrule(lr){6-7}
				& PSNR & SSIM & PSNR & SSIM & PSNR & SSIM \cr
				\midrule
				GMM \cite{Gaussian_mixture_model}    & 27.16 & 0.8982 & 13.04 & 0.4673 & 24.04 & 0.8675 \cr
				DSC \cite{DSC}      & 25.68 & 0.8751 & 13.17 & 0.4272 & 20.95 & 0.7530 \cr
				DDN \cite{detail_layer} & 33.01 & 0.9692 & 24.64 & 0.8489 & 24.68 & 0.8739 \cr
				RESCAN \cite{rescan}    & {36.39} & {0.9767} & 26.60 & {0.8974} & 24.09 & 0.8410 \cr
				DAF-Net \cite{hu2019} & 32.07 & 0.9641 & 24.65 & 0.8607 & 25.27 & 0.8895 \cr
				SPA-Net \cite{wang2019spatial} & 31.59 & 0.9652 & 23.04 & 0.8522 & 22.41 & 0.8382 \cr
				PReNet \cite{ren2019progressive} & 36.76 & 0.9796 & {28.08} & 0.8871 & 26.61& {0.9015} \cr
				MSPFN \cite{jiang2020multi} & 32.98 & 0.9693 & 27.38 & 0.8690 & 25.59& 0.8882 \cr
				MPRNet \cite{zamir2021multi} & 37.32 & 0.9810 & 28.32 & 0.9168 & 26.10& 0.8967 \cr
				AirNet \cite{li2022all} & 34.90 & 0.9660 & 25.48 & 0.8229 & 23.77& 0.8335 \cr
				DGUNet \cite{mou2022deep} & 38.25 & 0.9740 & 31.06 & 0.8972 & 26.22& 0.8893 \cr
				\midrule
				SIRR \cite{wei2019semi} & 35.46 & 0.9726 & 26.89 & 0.8463 & 24.36& 0.8698 \cr
				Syn2Real \cite{yasarla2020syn2real} & 34.39 & 0.9657 & 25.76 & 0.8370 & 24.24& 0.8667 \cr
				JRGR \cite{ye2021closing} & 31.98 & 0.9669 & 23.46 & 0.8449 & 22.16& 0.8228 \cr
				\midrule
				DRD-Net \cite{deng2020detail}    & 37.15 & 0.9873 & 28.16 & 0.9201& 26.32& 0.9018 \cr
				Semi-DRDNet & \textbf{40.66} & \textbf{0.9885} & \textbf{31.32} & \textbf{0.9240}& \textbf{27.73}& \textbf{0.9055} \cr
				\bottomrule
			\end{tabular}
		}
	\end{threeparttable}
\end{table*}

\subsection{Dataset}
\par \textbf{Synthetic Datasets:} 
For labeled synthetic images, we evaluate the performance of Semi-DRDNet on the commonly tested benchmark datasets: (1) Rain200L \cite{shuju} consists of 1800 training images and 200 test images with only one type of rain streaks; (2) Rain200H \cite{shuju} contains 1800 training images and 200 test images with five streak directions; (3) Rain800 \cite{zhang2019image} contains 700 training images and 100 testing images, which are synthesized following the guidelines in \cite{shuju}.

\textbf{Real-world Datasets:} For unlabeled real-world images, compared to DDN-SIRR \cite{wei2019semi} which contains only 147 real-world rainy images, we establish a new real-world rainy dataset called Real200. Real200 contains 2000 real-world rainy images (1800 training images and 200 testing images). Concretely, we first collect 436 real-world rainy images from \cite{wei2019semi,zhang2019image,yang2017deep,wu2019beyond,li2019single} and Google search with ``real rainy image". Then, we collect 100 real-world rainy videos and obtain 1564 rainy frames as real-world rainy images. It is noteworthy that we carefully check each image to ensure a large diversity of rain scenes.
Semi-DRDNet and many methods are trained in a semi-supervised manner. Following the protocols of \cite{yasarla2020syn2real,huang2021memory,wei2021semi}, we train them on three synthetic datasets (Rain200H, Rain200L, and Rain800) as labeled data and Real200 as unlabeled data, which are denoted by $\&$, such as Rain200H$\&$Real200, Rain200L$\&$Real200, and Rain800$\&$Real200.

\subsection{Training Details}
We implement Semi-DRDNet using Pytorch 1.6 on a system with 11th Gen Intel(R) Core(TM) i7-11700F CPU and Nvidia GeForce RTX 3090 GPU. During training, we set the depth of our network as 35, and utilize the non-linear activation PReLU \cite{PRelu}. 
For optimizing our network, we employ the Adam optimizer \cite{adam} with an initial learning rate of $1e^{-3}$. It takes about 150 epochs for our network to converge. When reaching 30, 50 and 80 epochs, the learning rate is
decayed by multiplying 0.2. All training images are cropped into 100$\times$100 patches with a batch size of 8. We set $\lambda_{unsup}$, $\lambda_{r}$ and $\lambda_{dual}$ to be $1$, $0.5$ and $0.5$, respectively.


\subsection{Comparison with the State-of-the-Arts}
\par \textbf{Baselines:} We compare Semi-DRDNet with 15 state-of-the-art deraining algorithms, including two prior-based methods, i.e., GMM \cite{Gaussian_mixture_model}, and DSC \cite{DSC}; 10 supervised deraining methods, i.e., DDN \cite{detail_layer}, RESCAN \cite{rescan}, DAF-Net \cite{hu2019}, SPA-Net \cite{wang2019spatial}, PReNet \cite{ren2019progressive}, MSPFN \cite{jiang2020multi}, DRD-Net \cite{deng2020detail}, MPRNet \cite{zamir2021multi}, AirNet \cite{li2022all} and DGUNet \cite{mou2022deep}; three semi-supervised deraining approaches, i.e., SIRR \cite{wei2019semi}, Syn2Real \cite{yasarla2020syn2real} and JRGR \cite{ye2021closing}. For the evaluations on synthetic and real-world images, all the supervised methods are directly trained on Rain200H, Rain200L, Rain800, and the semi-supervised ones are trained on Rain200H$\&$Real200, Rain200L$\&$Real200 and Rain800$\&$Real200. In the quantitative evaluation, Peak Signal-to-Noise Ratio (PSNR) and Structure Similarity Index (SSIM) are employed as the comparison criteria. More details of PSNR and SSIM can be seen in \cite{wang2004image}. Usually, a larger PSNR or SSIM indicate a better result.

\par
\begin{figure}[!t] \centering
	\includegraphics[width=1\linewidth]{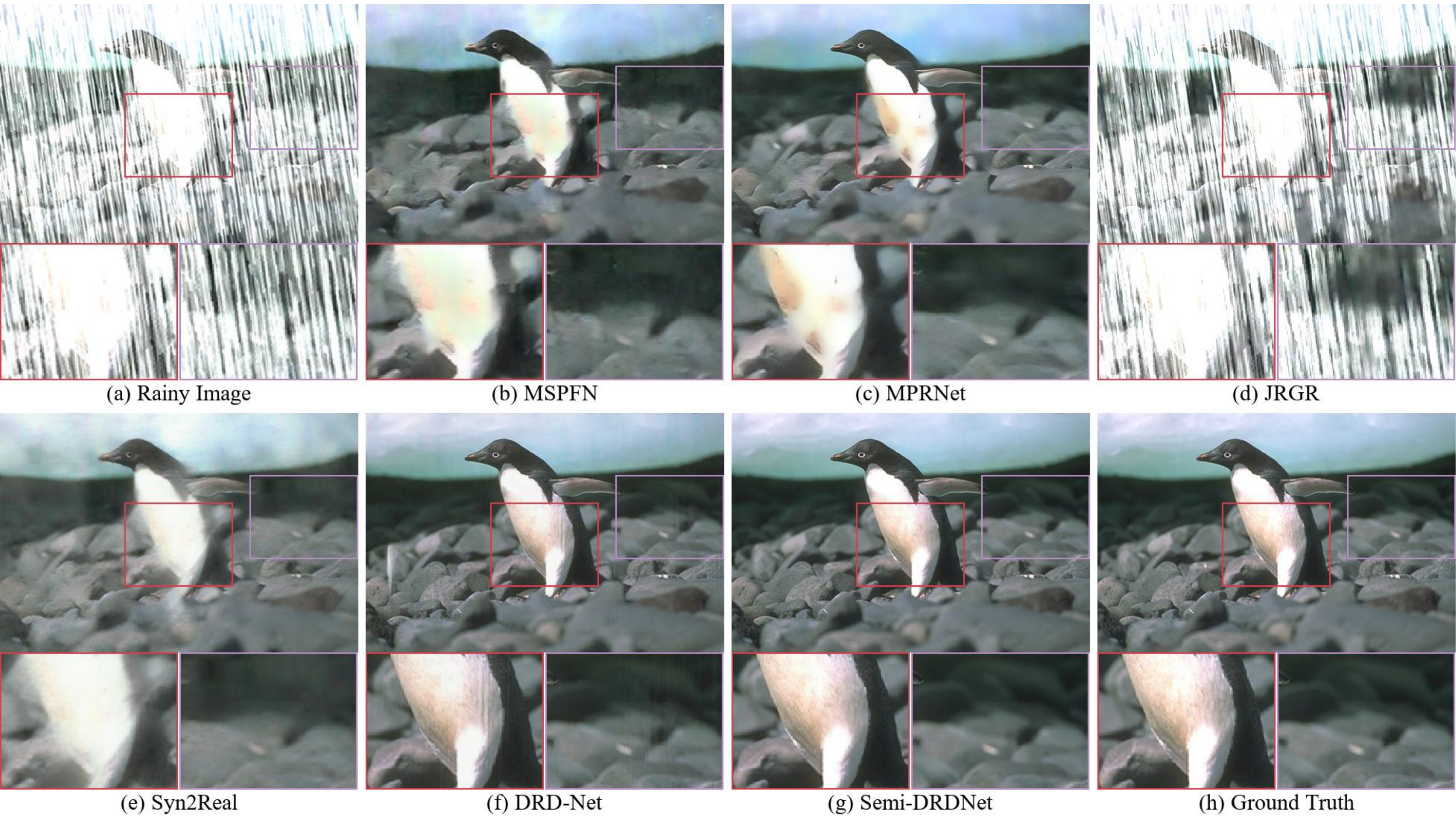}
	\caption{Test of removing large-scale rain streaks. Semi-DRDNet can avoid introducing additional artifacts while (b-f) MSPFN \cite{jiang2020multi}, MPRNet \cite{zamir2021multi}, JRGR \cite{ye2021closing}, Syn2Real \cite{yasarla2020syn2real} and DRD-Net \cite{deng2020detail} cannot do.}
	\label{fig:sys1}
\end{figure}

\begin{figure}[!t] \centering
	\includegraphics[width=1\linewidth]{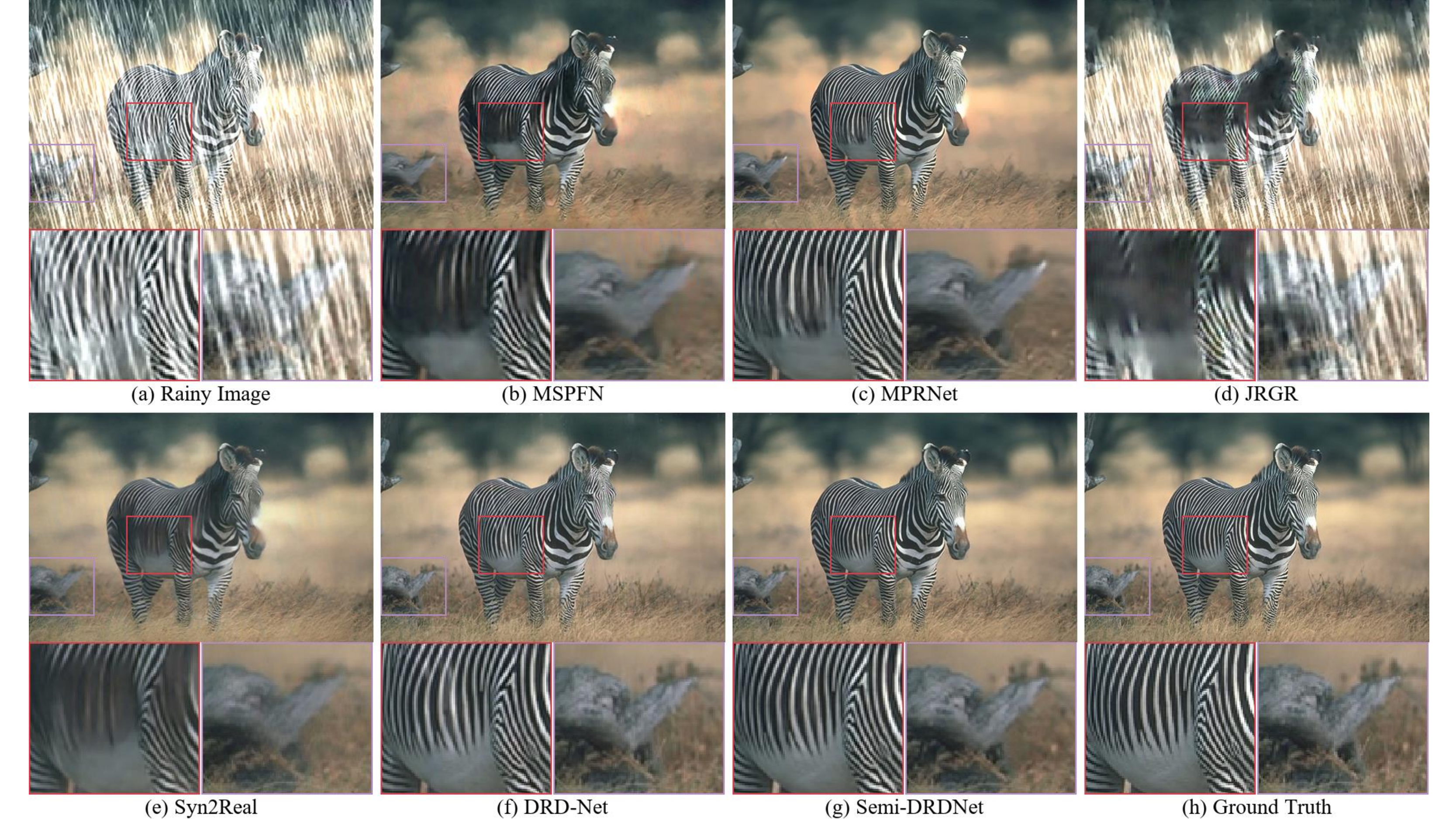}
		\caption{Test of preserving image details that have very similar patterns to rain streaks. Semi-DRDNet can distinguish image details and rain streaks well, and preserve these details better than (b-f) MSPFN \cite{jiang2020multi}, MPRNet \cite{zamir2021multi}, JRGR \cite{ye2021closing}, Syn2Real \cite{yasarla2020syn2real} and DRD-Net \cite{deng2020detail}.}
	\label{fig:sys2}
\end{figure}

 \textbf{Comparisons on the synthetic test sets:} 
Our method clearly outperforms all the deraining methods in terms of both PSNR and SSIM, as shown in Table \ref{Tab:compare}. 
Especially, Semi-DRDNet obtains more than 2.41 dB, 0.26 dB, and 1.51 dB PSNR gains on the test sets of Rain200L, Rain200H, and Rain800, compared with the latest supervised DGUNet \cite{mou2022deep}.
Besides, PSNR of our Semi-DRDNet gains over the semi-supervised Syn2Real \cite{yasarla2020syn2real} more than 6.27 dB, 5.56 dB, and 3.49 dB on Rain200L, Rain200H, and Rain800. Such large gains demonstrate the superiority of the proposed semi-supervised paradigm on synthesized rainy images.
Furthermore, compared to DRD-Net \cite{deng2020detail} (our conference version), Semi-DRDNet can take advantage of unlabeled real-world data to improve the results of image deraining, and obtain 3.51 dB, 3.16 dB, and 1.41 dB PSNR gains on Rain200L, Rain200H, and Rain800 respectively. 
We also show the visual results of different methods in Fig. \ref{fig:sys1} and Fig. \ref{fig:sys2}. It can be observed that Semi-DRDNet not only successfully removes the majority of rain streaks, but also effectively avoids image degradation caused by deraining, and better preserves texture details. Although most approaches can remove the rain streaks from the rainy image, the halo artifacts and color distortion have appeared after deraining.

\begin{figure*}[!ht] \centering
	\includegraphics[width=1\linewidth]{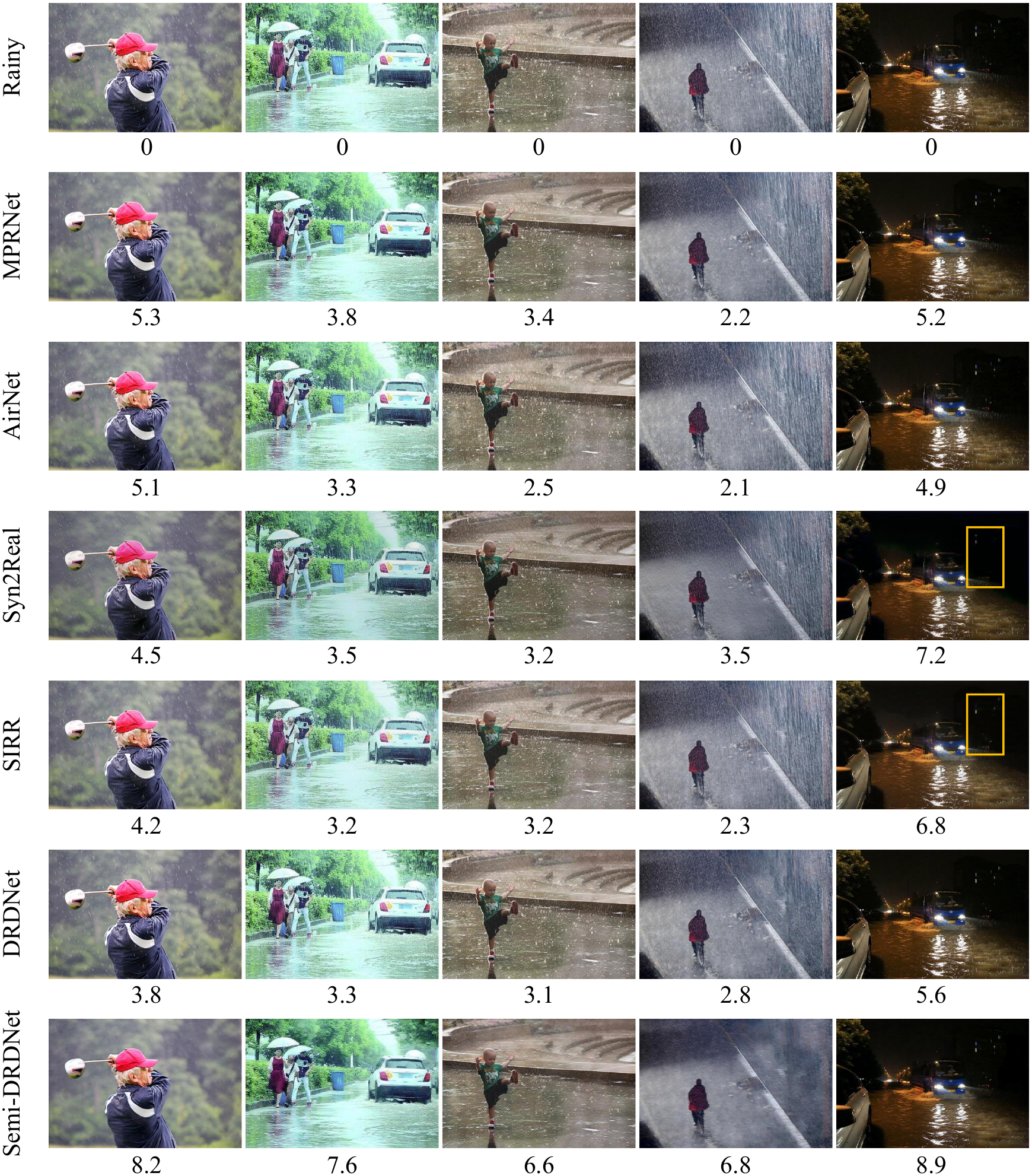}
	\caption{Image deraining results tested in Real200. The number below every image is the mean rating score in user study, which is provided in the supplementary material. Our method can well handle the rain streaks while preserving image details. Note that the yellow box represents the lost details.}
	\label{fig:realrain_result}
\end{figure*}

\textbf{Comparisons on real-world rainy images:} 
We evaluate Semi-DRDNet on the real-world test set of Real200 as shown in Fig. \ref{fig:realrain_result}. 
We also conduct a user study for subjective assessment, which is provided in the supplementary material.
Our method can effectively remove rain streaks of various appearances (e.g., small rain streaks and large rain streaks in the first and second columns of Fig. \ref{fig:realrain_result}).
The third and fourth columns of Fig. \ref{fig:realrain_result} show the heavy rain scenes where all compared methods fail to remove dense rain accumulation in heavy rainy images. Comparatively, our method is more successful to handle heavy rain streaks and achieve relatively better deraining results.
Further, the fifth column of Fig. \ref{fig:realrain_result} represents the nighttime rainy image, where without global uniform atmospheric light, the deraining results of most approaches become darker and some details turn invisible, while our Semi-DRDNet still achieves better deraining results while preserving their details.
Compared to DRD-Net \cite{deng2020detail}, our Semi-DRDNet achieves superior deraining performance on both synthetic and real-world rainy images. Such significant improvement demonstrates that the semi-supervised learning paradigm with DualSCRNet significantly boosts the performance on a large diversity of real-world rainy images.

\begin{table*}[!t]
	\centering
\footnotesize
	\caption{Quantitative comparison between our Semi-DRDNet and other network architectures on the test sets of Rain200L, Rain200H and Rain800.}
	\label{Tab:ablation_1}
	\begin{threeparttable}
		\centering
		\setlength{\tabcolsep}{4.5mm}{
			\begin{tabular}{ccccccccc}
				\toprule
				Dataset & Metrics & BL & BL+SE & BL+SE+DB & BL+SE+RB & BL+SE+SDCAB&Semi-DRDNet \cr
				\toprule
				\multirow{2}{*}{Rain200L}
				& PSNR & 35.69 & 36.25 & 36.93 & {37.16} & 37.52&\textbf{40.66}  \cr
				& SSIM & 0.9782 & 0.9796 & 0.9804 & {0.9876} & 0.9880&\textbf{0.9885}  \cr
				\multirow{2}{*}{Rain200H}
				& PSNR & 26.33 & 26.79 & 27.45 & 27.32 & 28.32&\textbf{31.32}  \cr
				& SSIM & 0.8323 & 0.8441 & 0.9194 & 0.9086 & 0.9221&\textbf{0.9240}  \cr
				\multirow{2}{*}{Rain800}
				& PSNR & 25.94 & 26.35 & 26.39 & 26.45 & 26.56&\textbf{27.73}  \cr
				& SSIM & 0.8132 & 0.8217 & 0.8945 &0.8978 & 0.9024&\textbf{0.9055}  \cr
				\bottomrule
			\end{tabular}
		}
	\end{threeparttable}
\end{table*}

\begin{figure}[!t] \centering
	\includegraphics[width=1\linewidth]{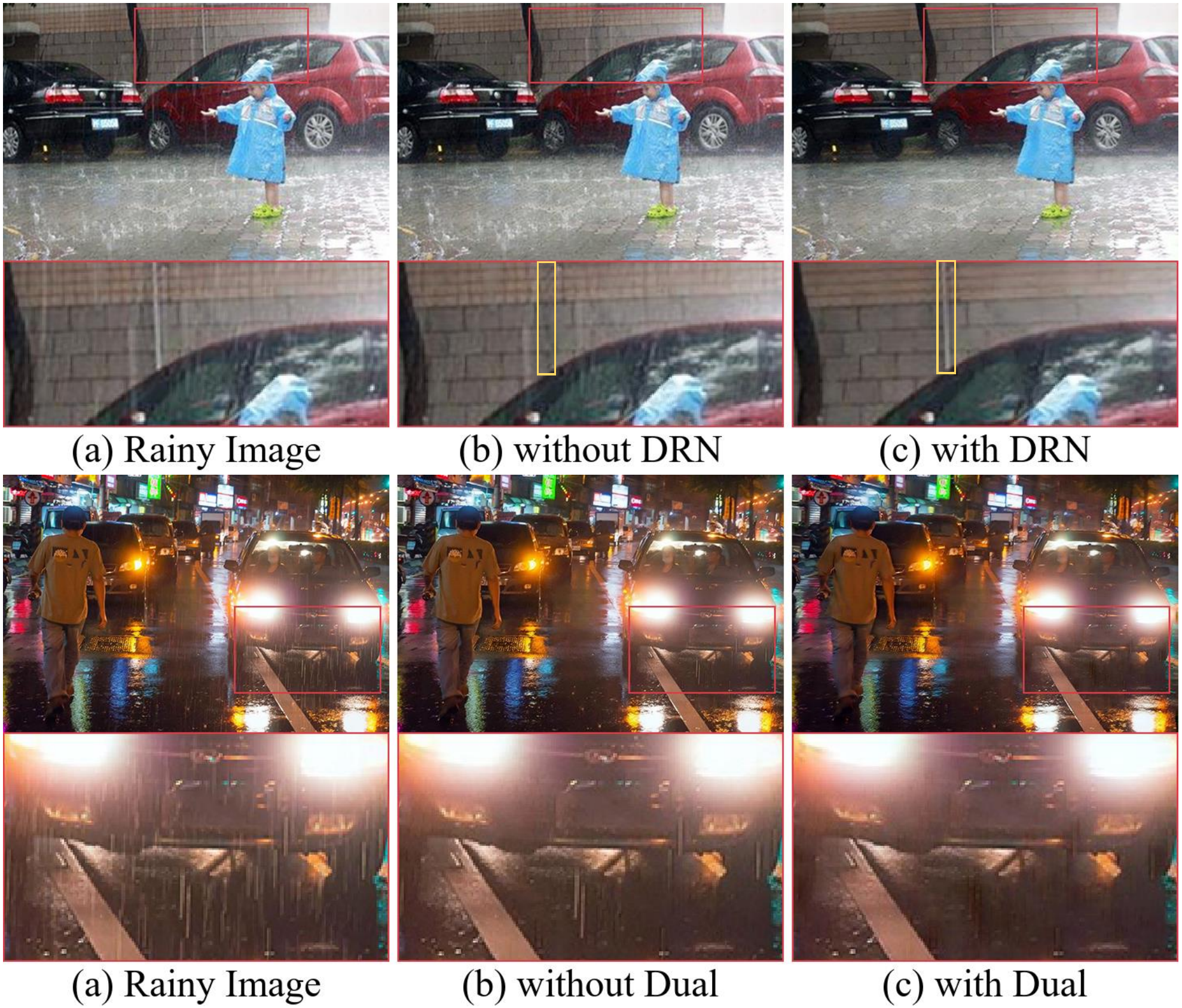}
	\caption{
	Validation of our DRN and DualSCRNet in real-world rainy images. (b) denotes Semi-DRDNet without DRN and DualSCRNet, and (c) denotes Semi-DRDNet with DRN and DualSCRNet. Note that the yellow box represents the lost details.
	}
	\label{fig:ab_real1}
\end{figure}

\subsection{Ablation Study}
We conduct ablation experiments to gain insight into the respective roles of different components and loss functions. 

\par\textbf{Ablation Study on Different Components:} To explore the effectiveness of our Semi-DRDNet, it is necessary to decompose its full scheme into different parts and even replace the network architecture for the ablation study.

\begin{itemize}
	\item \textbf{BL:} Baseline (BL) indicates the residual network without the SE operation, which learns a function that maps the rainy images to the rain streaks.
	
	\item \textbf{BL+SE:} Adding the SE operation to the baseline.
	
	\item \textbf{BL+SE+DB:} Employing two sub-networks for image deraining. One network is the rain residual network (BL+SE), and the another is detail repair network based on the direct block (DB, see in Fig.  \ref{fig:different connect}(a)).
	
	\item \textbf{BL+SE+RB:} DB is replaced with residual block (RB) in the detail repair network.
	
	\item \textbf{BL+SE+SDCAB:} Comprising the rain residual network (BL+SE) and the detail repair network based on the proposed structure detail context aggregation block (SDCAB). Although this model has the same network structure as DRD-Net\cite{deng2020detail} (our conference version), it trains in a semi-supervised manner with different loss functions.
	\item \textbf{BL+SE+SDCAB+DualSCRNet:} Semi-DRDNet comprises the rain residual network (BL+SE), the detail repair network based on SDCAB, and  DualSCRNet.
\end{itemize}

\begin{table}[!t]
	\centering
\footnotesize
	\caption{Ablation study on different settings of our method in Rain200H. M is the number of feature maps and D is the total depth of our network.}
	\label{Tab:ablation_2}
	\begin{threeparttable}
		\centering
		\small
		\setlength{\tabcolsep}{3mm}{
			\begin{tabular}{ccccc}
				\toprule
				& Metrics & M = 16 & M = 32 & M = 64  \cr
				\toprule
				\multirow{2}{*}{D = 8+3}
				& PSNR & 27.67 & 28.82 & 29.96 \cr
				& SSIM & 0.9138 & 0.9164 &0.9188 \cr
				\multirow{2}{*}{D = 12+3}
				& PSNR & 28.70 & 29.91 & 30.67  \cr
				& SSIM & 0.9156 & 0.9177 & 0.9203  \cr
				\multirow{2}{*}{D = 16+3}
				& PSNR & 29.38 & 30.13 & \textbf{31.32}  \cr
				& SSIM & 0.9175 & 0.9199 & \textbf{0.9240}  \cr
				\bottomrule
			\end{tabular}
		}
	\end{threeparttable}
\end{table}

			

\begin{figure}[!t] \centering
	\includegraphics[width=1\linewidth]{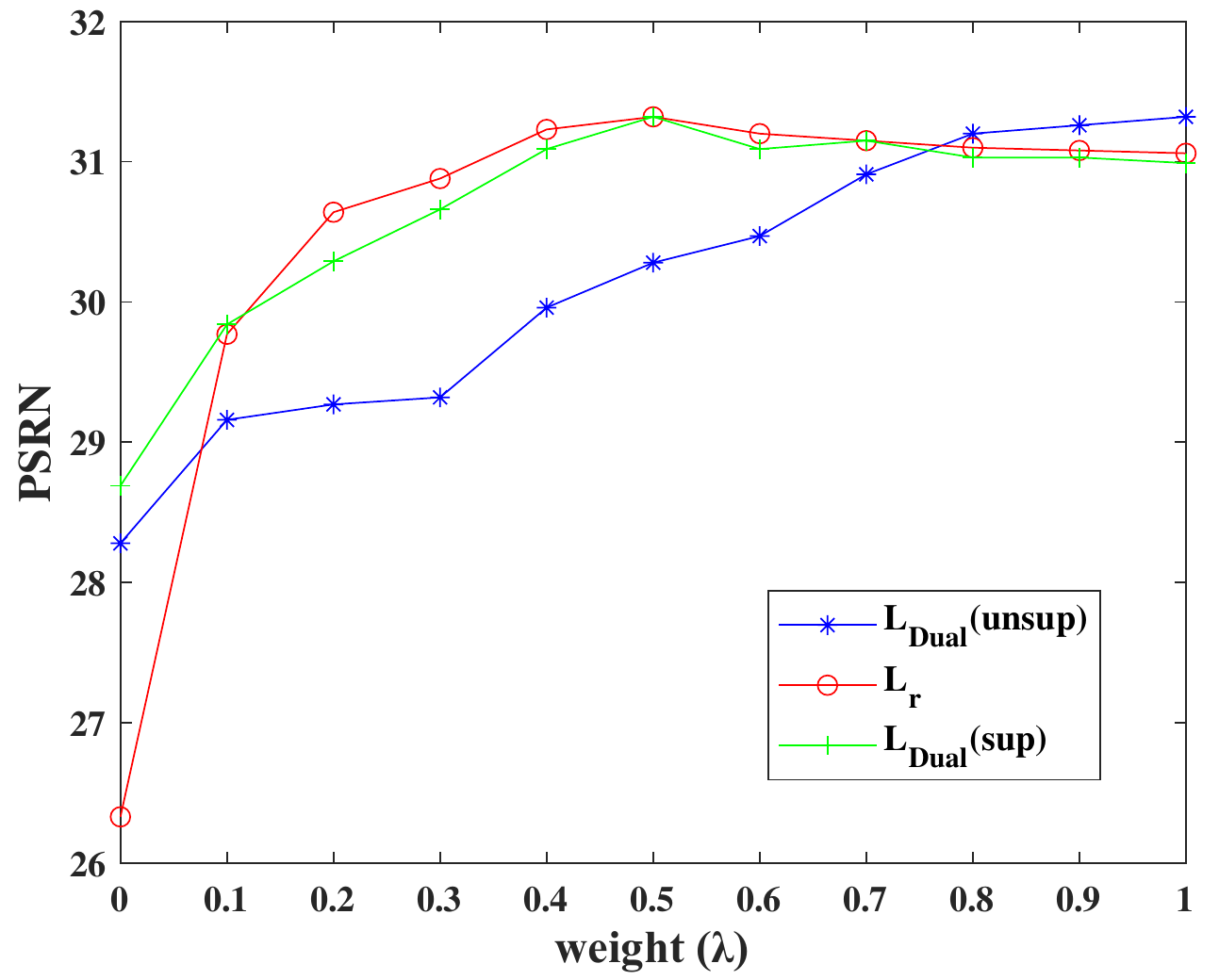}
	\caption{Effectiveness of different loss functions of Semi-DRDNet.}
	\label{fig:loss}
\end{figure}

\textbf{Effect of SE, SDCAB and DualSCRNet:} To validate the necessity of the structure in Figs. \ref{fig:network framework} and \ref{fig:different connect}, we show the results in Table \ref{Tab:ablation_1} and Fig. \ref{fig:ab_real1}. 
It is found that: (i) The performance of deraining without the SE operation suffers from slight degradation. This certifies the necessity of the SE operation from another side. 
(ii) The performance of deraining without the detail recovery network suffers from image detail blurring in the real-world images (Fig. \ref{fig:ab_real1}), which proves the necessity of DRN to find the lost details. 
(iii) In order to evaluate the effectiveness of SDCAB, we compare our network with other connection style blocks, including the direct block (DB), and the residual block (RB) which has been used in DDN \cite{detail_layer}. For fair comparisons, we replace SDCAB with DB and RB respectively, the result (shown in Table \ref{Tab:ablation_1}) certifies that SDCAB is essential to detail-recovery image deraining. 
(iv) Moreover, the full scheme of Semi-DRDNet (BL+SE+SDCAB+DualSCRNet) outperforms other architectures both quantitatively and qualitatively, which certifies that DualSCRNet can constrain the deraining network to approximate the clean images and move away from the real-world rainy images, thus benefiting real-world rain removal. 
Since both DRN and DualSCRNet can be easily attached to existing deraining networks to boost their performance,
we provide further analysis of them in the supplementary material.

\par \textbf{Effect of Parameter Settings:} Results under different parameter settings of Semi-DRDNet can be found in Table \ref{Tab:ablation_2}. We have discussed the effects of the number of feature maps and SDCAB or the rain residual blocks. The table shows that more parameters lead to higher performance.

\par \textbf{Effect of Loss Functions:} 
We evaluate the deraining performance by considering different combinations of loss functions in Rain200H (see Fig. \ref{fig:loss}): 
i) The introduction of the rain residual loss and two contrastive losses greatly improves the model performance. 
ii) The introduction of unsupervised loss brings significant improvements on the deraining accuracy by using unpaired real-world data. Especially, compared to the complex unsupervised loss functions of existing semi-supervised deraining, we only adopt the unsupervised contrastive loss as our unsupervised loss function.
iii) The balanced weights between different losses are chosen experimentally, which optimizes our network effectively.

\section{Conclusion}
In this paper, we propose a semi-supervised detail-recovery image deraining network (Semi-DRDNet) for semi-supervised image deraining. It alleviates two types of domain gaps between synthetic and real-world rainy images to remove real-world rain and recover correct backgrounds.
Semi-DRDNet consists of three sub-networks. First, a rain residual network is designed to remove the rain streaks from the rainy images. Second, a detail repair network is proposed to find back the details of derained images. Third, a novel dual sample-augmented contrastive regularization network is developed to bridge two types of domain gaps of real-world image deraining, thus enhancing the real-world image deraining capacity. Qualitative and quantitative experiments indicate that our method outperforms the state-of-the-art supervised and semi-supervised approaches in terms of removing the rain streaks and recovering the image details.



\bibliographystyle{IEEEtran}
\bibliography{egbib}
\appendix

\subsection{User study on real-world rainy images} 
\begin{figure}[!t] \centering
	\includegraphics[width=1\linewidth]{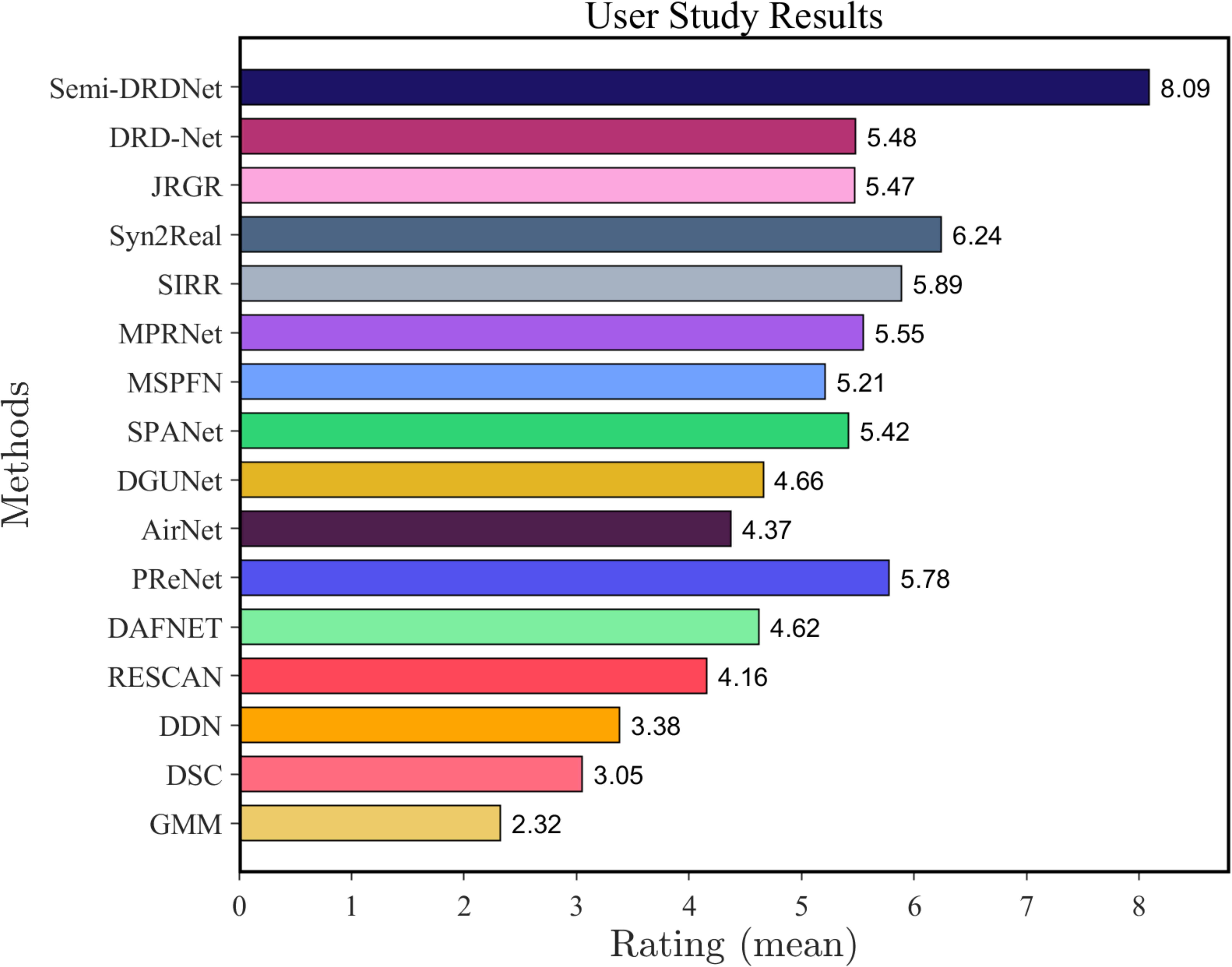}
	\caption{User study results. Mean ratings (from 1 (bad) to 10 (good)) given by the participants in Real200.}
	\label{Tab:user_study}
\end{figure}

Evaluation on real-world data that lacks the ground truths is commonly difficult and subjective. We conduct
a user study for subjective assessment: (i) To ensure fairness, we randomly choose 50 images covering different scene types from the test set of Real200. (ii) We recruit 100 volunteers to rank each derained image with the score from 1 (the worst) to 10 (the best): 37 females and 63 males, aged 16 to 30 with a mean of 25.5. (iii) We present rainy images and derained images to each volunteer in a random order, and do not tell them which approach the derained image is generated by, then ask each of them to rate how the quality of the derained image is on a scale from 1 to 10. (iv) We obtain 5000 ratings (100 volunteers × 50 images per category) altogether per category: our Semi-DRDNet and the other approaches. Fig. \ref{Tab:user_study} reports the results, showing that our Semi-DRDNet has more pleasing derained images than the others. At the end of the user study, some participants report that for the majority of our derained images, they see no water splashing on the ground like the clean photos.

\begin{figure}[!t] \centering
	\includegraphics[width=1.0\linewidth]{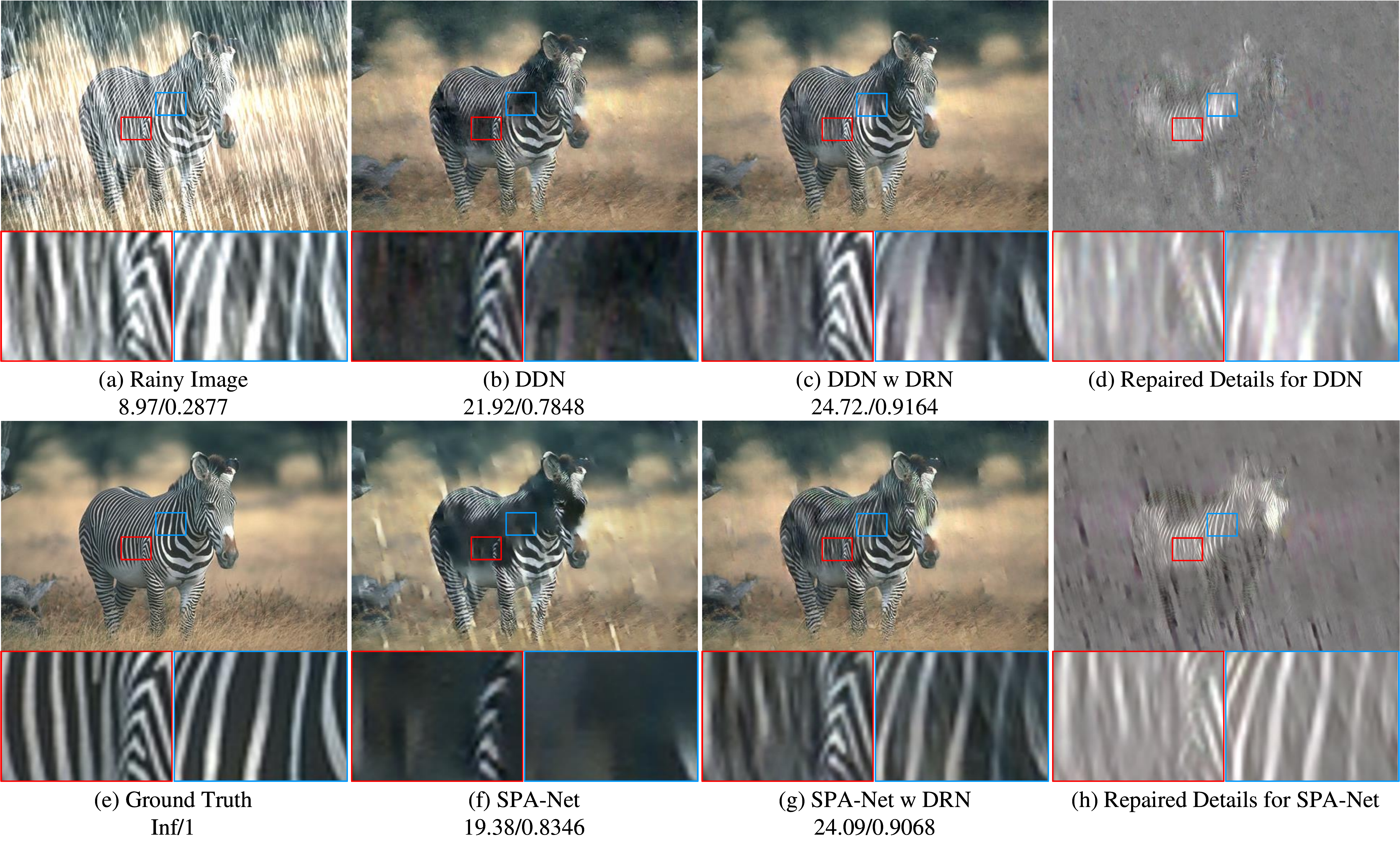}
	\caption{Validation of our DRN on Zebra image to repair image details that are similar to rain streaks. From (b)-(h): the deraining results of (b) DDN \cite{detail_layer}, (c) DDN with Detail Repair Network, (f) SPA-Net \cite{wang2019spatial}, (g) SPA-Net with Detail Repair Network, (d) and (h) are the repaired details of DDN and SPA-Net, and (e) GT.}
	\label{fig:Zebra}
\end{figure}

\begin{figure}[!t] \centering
	\includegraphics[width=1.0\linewidth]{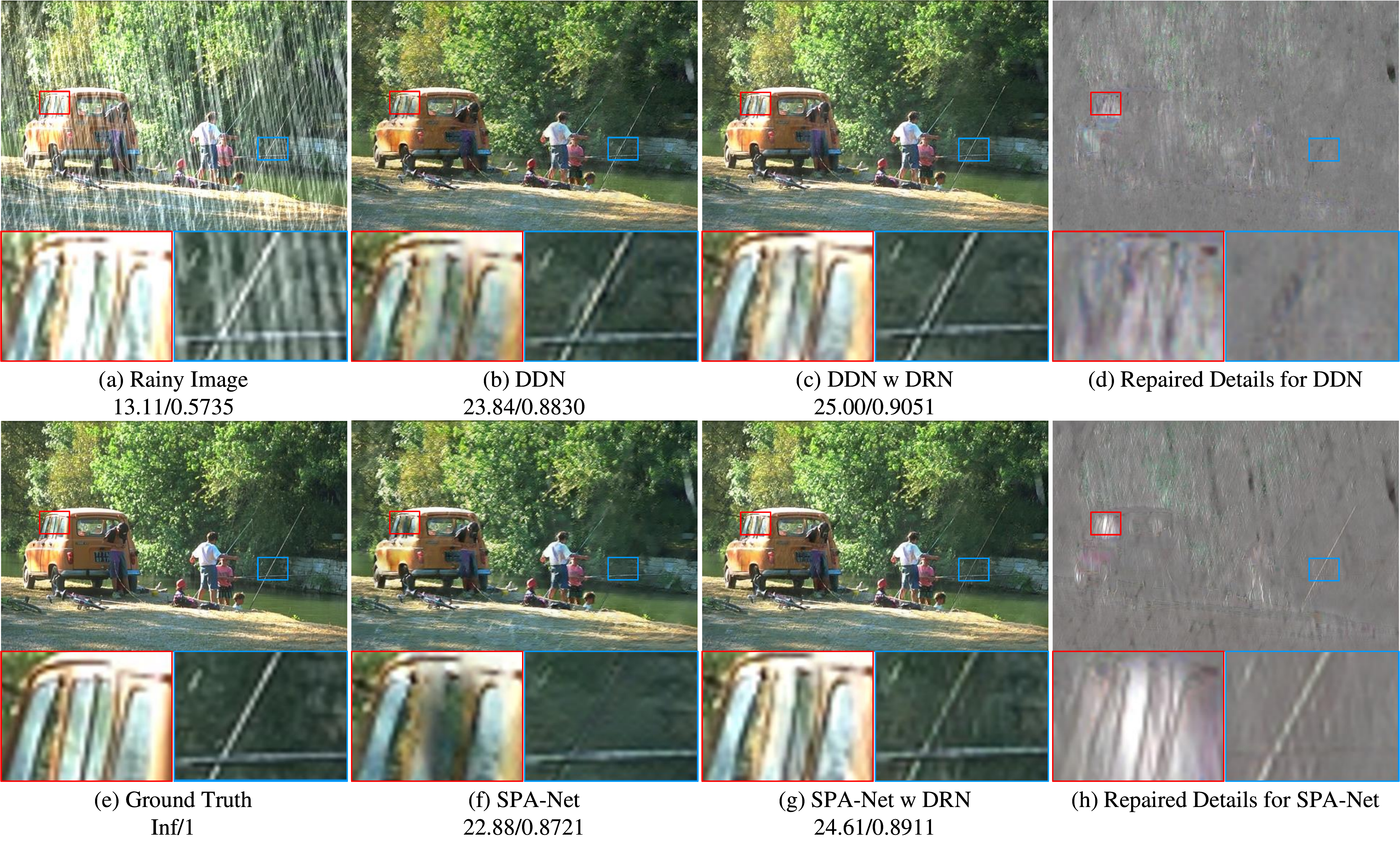}
	\caption{Validation of our DRN on Car image to repair very weak details that are easily submerged in rain streaks. From (a)-(h): the deraining results of (b) DDN \cite{detail_layer}, (c) DDN with Detail Repair Network, (f) SPA-Net \cite{wang2019spatial}, (g) SPA-Net with Detail Repair Network, (d) and (h) are the repaired details of DDN and SPA-Net, and (e) GT.}
	\label{fig:Car}
\end{figure}

\begin{table}[!t]
	\centering
	\caption{Evaluations of combinations of two deraining models and our DDN, i.e., DDN \cite{detail_layer}/SPA \cite{wang2019spatial} + DRN.} 
	\label{Tab:ddn_with_rpn}
	\begin{threeparttable}
		
		\footnotesize
		\centering
		\setlength{\tabcolsep}{1.5mm}{
			\begin{tabular}{cccccc}
				\toprule
				Datasets&  Metrics& DDN& DDN w/ DRN& SPA& SPA w/ DRN  \cr
				\toprule
				\multirow{2}{*}{Rain200H}
				& PSNR & 24.64 & 25.92 & 23.04 & 25.68  \cr
				& Time & 0.03s & 0.15s & 0.06s & 0.45s  \cr
				\multirow{2}{*}{Rain800}
				& PSNR & 24.04 & 25.13 & 22.41 & 25.67 \cr
				& Time & 0.05s & 0.14s & 0.26s & 0.35s \cr
				\bottomrule
			\end{tabular}
		}
	\end{threeparttable}
\end{table}
\subsection{Analysis of DRN} 
Existing learning-based deraining methods resort to delicate network design to meet the challenging goal of removing rain streaks but retaining details of similar properties. In contrast, our Semi-DRDNet decomposes this conflicting task into `\textbf{remove}' and `\textbf{repair}' by two parallel sub-networks, which share the same input and collaborate to spit a high-fidelity output. Apparently, the choice of the rain removal part is not unique, the detail recovery sub-network can be easily attached to existing deraining networks to boost their performance. 
\begin{figure}[!t]
	\centering
	\includegraphics[width=1\linewidth]{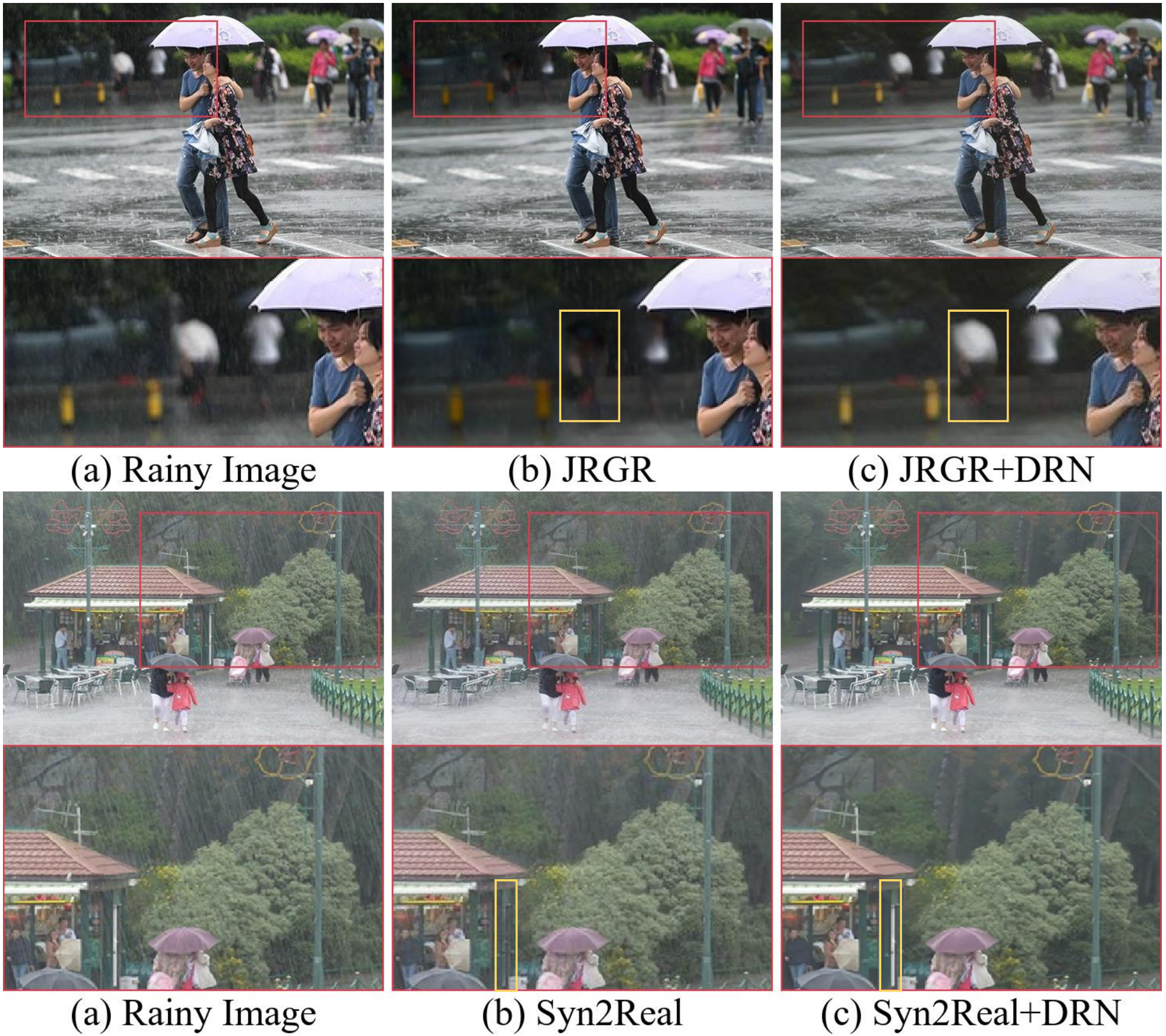}
	\caption{Our detail-recovery network facilitates other semi-supervised methods to find back the lost details during deraining. Note that all models are trained on Rain200H$\&$Real200.}
	\label{fig:drn}
\end{figure}

\textbf{Detail Recovery for Synthetic Images.} Taking DDN \cite{detail_layer} and SPA \cite{wang2019spatial} as examples, we experiment with two parallel networks consisting of our detail repair network and their deraining networks. For fair comparisons, we keep most parameters from the original DDN and SPA untouched. The depth and the number of feature channels of the detail repair network are set as 24 and 16 respectively. We randomly select 20 image patches with the size of $64 \times 64$ to train the networks. We compare  PSNR and the running time of deraining methods with and without our detail recovery mechanism on different datasets. From Table \ref{Tab:ddn_with_rpn} we observe that attaching our detail repair network considerably improves PSNR while sacrificing negligible time efficiency. 
Figs. \ref{fig:Zebra} and \ref{fig:Car} show that deraining networks tend to lose details which share similar properties with rain streaks, such as the zebra-stripe in Fig. \ref{fig:Zebra} and the fishing rod in  Fig. \ref{fig:Car}. We demonstrate that by simply incorporating our detail repair network, these lost details can be easily added back to restore the high-quality rain-free images.

\textbf{Detail Recovery for Real-world Images.} To our knowledge, recent semi-supervised deraining paradigms pay little attention to detail recovery yet. To certify that the detail recovery is also essential to real-world image deraining, we incorporate DRN into the semi-supervised deraining paradigms \cite{ye2021closing} and \cite{yasarla2020syn2real}. It is noteworthy that JRGR includes four deraining sub-networks, thus we add four additional parallel DRNs to these sub-networks for the joint training of JRGR.  From Fig. \ref{fig:drn} we can observe that DRN can also effectively find back the lost details during the semi-supervised deraining process, and obtain better deraining performance on real-world images. Thus, it is reasonable to view rain removal and detail recovery as two separate tasks, so that each part could be specialized rather than traded off.

\begin{table}[!t]
	\centering
	\caption{Qualitative comparisons between SDCAB and MSARR. SDCAB and MSARR indicate DRN based on SDCAB and MSARR, respectively.}
	\label{Tab:masraa}
	\begin{threeparttable}
		
		\small
		\centering
		\setlength{\tabcolsep}{1.5mm}{
			\begin{tabular}{cccc}
				\toprule
				Datasets&  Metrics&  SDCAB &  MSARR\cr
				\toprule
				\multirow{2}{*}{Rain200H}
				& PSNR & 31.32 & 28.92    \cr
				& Time & 0.48s & 0.42s   \cr
				\multirow{2}{*}{Rain800}
				& PSNR & 27.73 & 26.29   \cr
				& Time & 0.64s & 0.58s   \cr
				\bottomrule
			\end{tabular}
		}
	\end{threeparttable}
\end{table}

\begin{table*}[tp]
	\small
	\centering	
	\caption{The detailed architecture of detail repair network. }
	\label{Tab:struct}
	\resizebox{\textwidth}{!}{
		\setlength{\tabcolsep}{4mm}{
			\begin{tabular}{lccccccccc}
				\hline
				\textbf{Layer} & 0 & 1 & 2  & $\dots$ & d  & $\dots$ & 16 & 17  &18  \\ \hline
				\textit{Convolution}      & $3 \times 3$    & $3 \times 3$     & $3 \times 3$     & $\dots$    & $3 \times 3$   & $\dots$     & $3 \times 3$    & $3 \times 3$    & $3 \times 3$   \\
				\textit{SDCAB}      & No    & Yes     & Yes     & $\dots$    & Yes   & $\dots$    & Yes   & No    & No   \\
				\textit{Dilation}         & 1         & 7       & 7  & $\dots$  & 7  & $\dots$    & 7         & 1   & 1\\
				\textit{Receptive field}  & $3 \times 3$    & $17 \times 17$    & $31 \times 31$   & $\dots$  & $(d-1) \times 14+17$  & $\dots$     & $227 \times 227$       & $229 \times 229$         & $231 \times 231$   \\ \hline
			\end{tabular}
		}
	}
\end{table*}
\textbf{SDCAB vs MSARR.} 
We train our detail repair network based on MSARR and observe that the performance drops from 31.32 dB to 28.64 dB on the test set of Rain200H (see Table \ref{Tab:masraa}), compared to DRN based on SDCAB. This shows that SDCAB can be used for designing a better detail recovery sub-network. The detailed structure of the detail repair network is presented in Table \ref{Tab:struct}, illustrating how the receptive field grows by applying the SDCAB block with multi-scale dilations.

\begin{table}[!t]
	\centering
	\caption{Quantitative comparison between our DualSCRNet and the standard contrastive learning module on the test sets of Rain200L, Rain200H and Rain800. Stand, Aug (pos), Aug (neg) and DualSCRNet represent the standard contrastive learning module, the positive sample-augmented strategy, the negative sample-augmented strategy and our DualSCRNet, respectively.}
	\label{Tab:aug_ab}
	\begin{threeparttable}
		
		\footnotesize
		\centering
		\setlength{\tabcolsep}{1.5mm}{
			\begin{tabular}{cccccc}
				\toprule
				Datasets&  Metrics& Stand&Aug (pos)&Aug (neg)& DualSCRNet\cr
				\toprule
				Rain200L
				& PSNR & 38.05 & 39.86 & 39.15 & 40.66 \cr
				Rain200H
				& PSNR & 28.96 & 30.02 & 29.76 & 31.32  \cr
				Rain800
				& PSNR & 26.82 & 27.21 & 27.18 & 27.73 \cr
				\bottomrule
			\end{tabular}
		}
	\end{threeparttable}
\end{table}

\subsection{Analysis of DualSCRNet} 
Many deraining methods overlook two types of domain gaps that exist between synthetic and real-world rainy images. To address it, a dual sample-augmented contrastive regularization network (DualSCRNet) is designed to build efficient contrastive constraints for both rain streaks and clean
backgrounds. Similar to DRN, our DualSCRNet can be easily attached to existing deraining networks to improve their deraining performance.

\textbf{Effect of Sample-augmented Strategies:} Unlike standard contrastive learning, we design two core sample-augmented strategies to produce a more discriminative visual mapping. We show the quantitative results
in Table \ref{Tab:aug_ab} to validate their necessity, showing that the two sample-augmented strategies effectively improve the deraining quality.
\begin{table}[!t]
	\centering
	\caption{Evaluations of combinations of two deraining models and our DualSCRNet, i.e., JRGR \cite{ye2021closing}/Syn2Real \cite{yasarla2020syn2real} + DualSCRNet.}
	\label{Tab:Dual-semi}
	\begin{threeparttable}
		
		\footnotesize
		\centering
		\setlength{\tabcolsep}{1.0mm}{
			\begin{tabular}{cccccc}
				\toprule
				Datasets&  Metrics& JRGR& JRGR w/ Dual& Syn2Real& Syn2Real w/ Dual  \cr
				\toprule
				\multirow{2}{*}{Rain200L}
				& PSNR & 31.98 & 36.92 & 34.39 & 37.36  \cr
				& SSIM & 0.9669 & 0.9691 & 0.9657 & 0.9825  \cr
				\multirow{2}{*}{Rain200H}
				& PSNR & 23.46 & 25.57 & 25.76 & 27.88 \cr
				& SSIM & 0.8449 & 0.8660 & 0.8370 & 0.8755 \cr
				\bottomrule
			\end{tabular}
		}
	\end{threeparttable}
\end{table}
\begin{figure}[!t]
	\centering
	\includegraphics[width=0.9\linewidth]{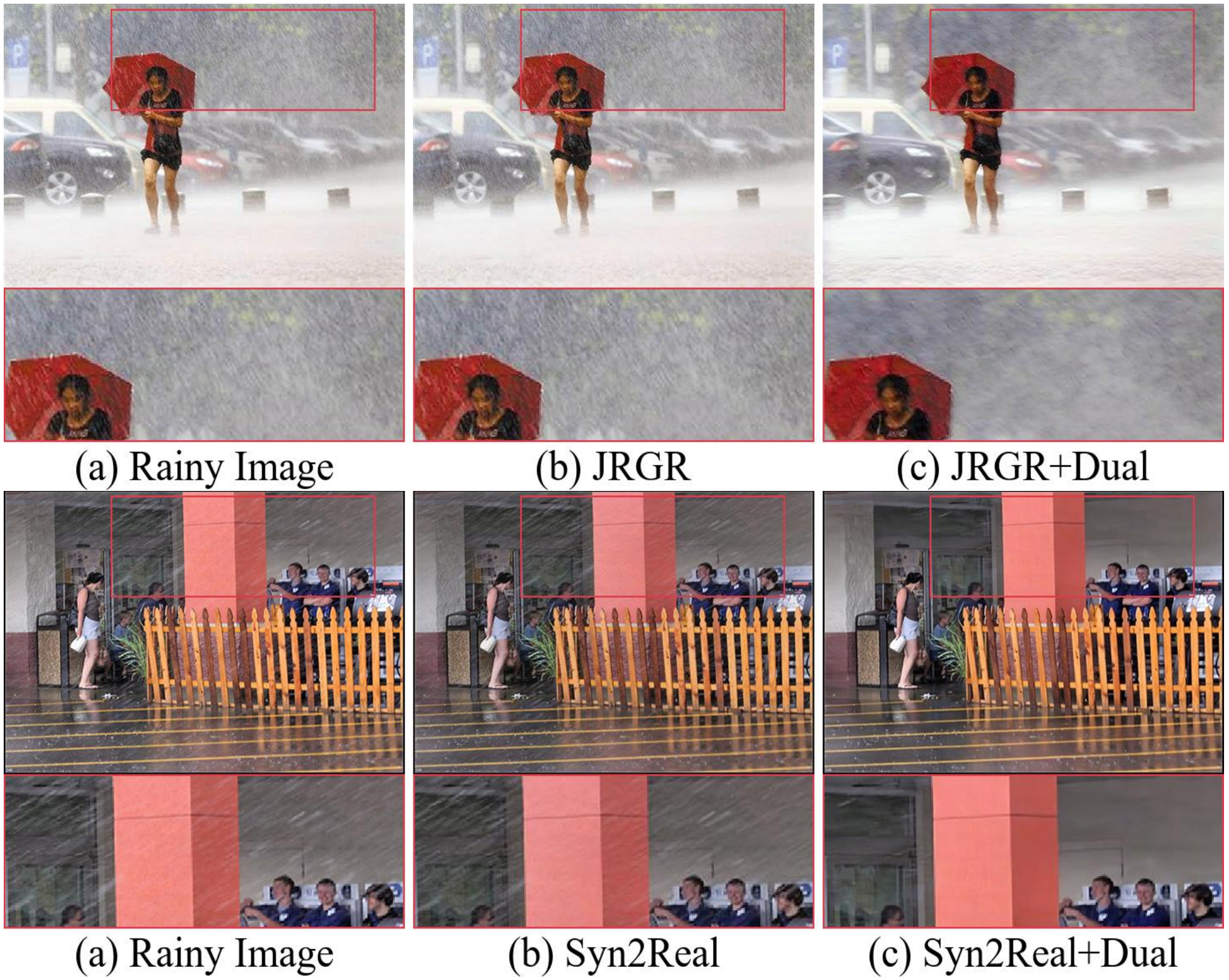}
	\caption{Our DualSCRNet improves the generalization ability of other semi-supervised methods by providing abundant image details. Note that all models are trained on Rain200H$\&$Real200. 
	}
	\label{fig:dual_incop}
\end{figure}

\textbf{DualSCRNet for Semi-supervised Image Deraining:} To verify the effectiveness of DualSCRNet for semi-supervised image deraining, we also incorporate DualSCRNet into two semi-supervised deraining methods, i.e., JRGR \cite{ye2021closing} and Syn2Real \cite{yasarla2020syn2real}. As shown in Table \ref{Tab:Dual-semi}, JRGR \cite{ye2021closing} and Syn2Real \cite{yasarla2020syn2real} have higher PSNR and SSIM than their original versions. The improvement comes from our DualSCRNet, which proves the importance of building efficient contrastive constraints. Moreover, as shown in Fig. \ref{fig:dual_incop}, DualSCRNet can improve the generalization ability of existing semi-supervised deraining paradigms and help them to obtain better real-world deraining results, especially in real-world images with heavy rain.

\begin{figure}[!t] \centering
	\includegraphics[width=1\linewidth]{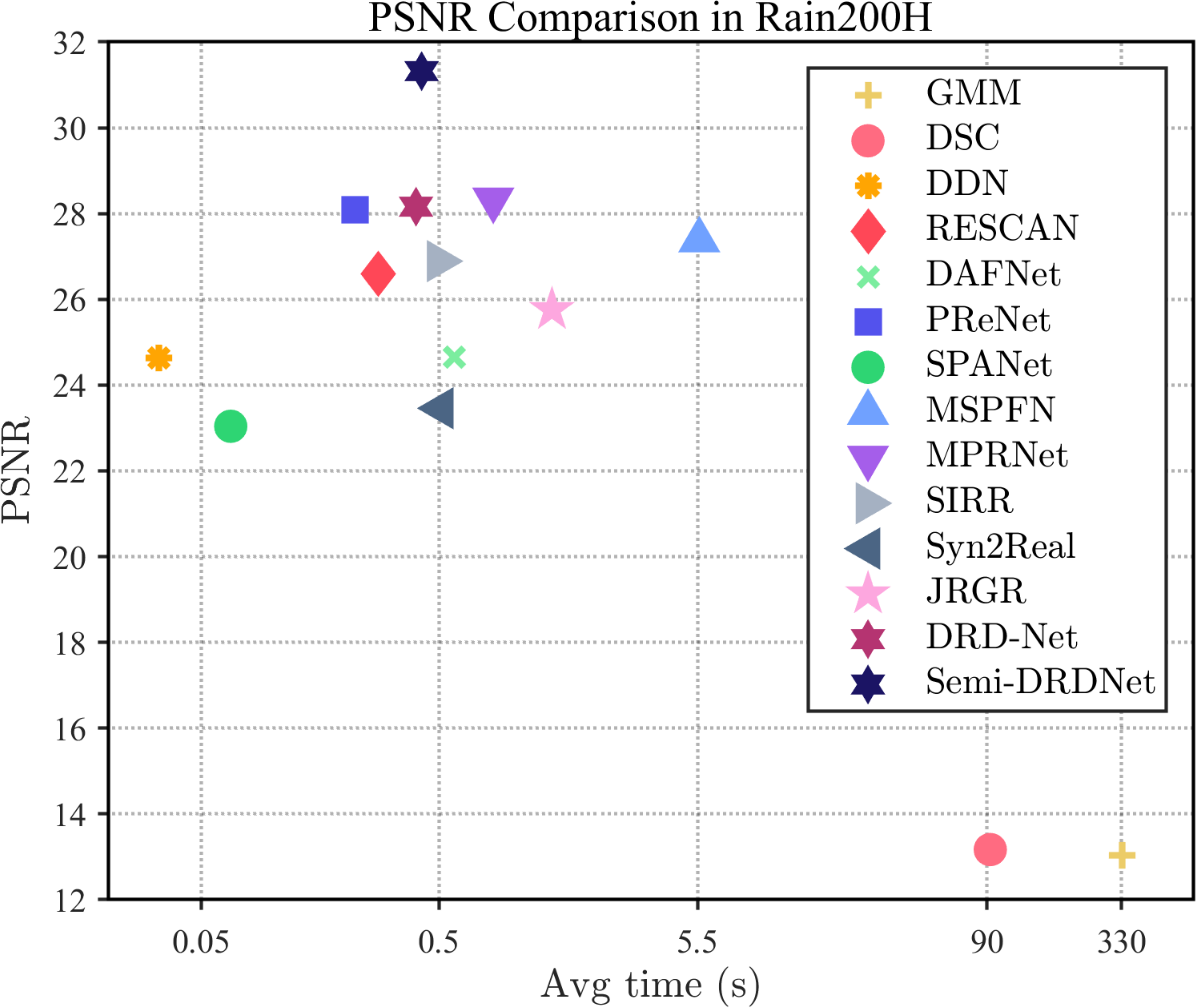}
	\caption{Averaged time and performance of different methods.}
	\label{Tab:time}
\end{figure}

\begin{figure*}!t]
	\centering
\footnotesize
	\subfigure[]{
		\includegraphics[width=2.1in, height=1.6in]{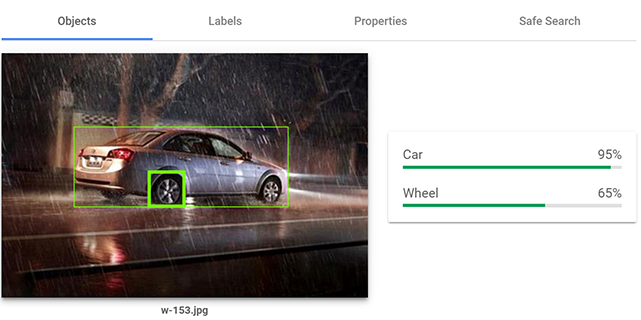}
	}
	\subfigure[]{
		\includegraphics[width=2.1in, height=1.6in]{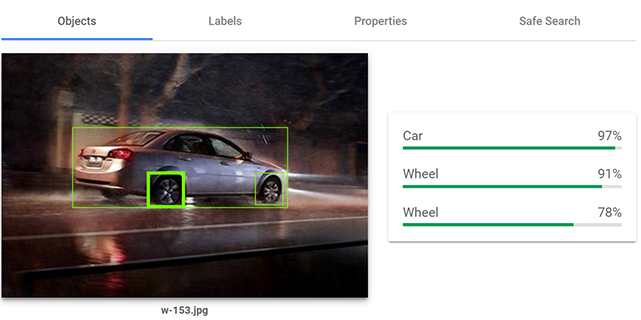}
	}
	\subfigure[]{
		\includegraphics[width=2.1in, height=1.6in]{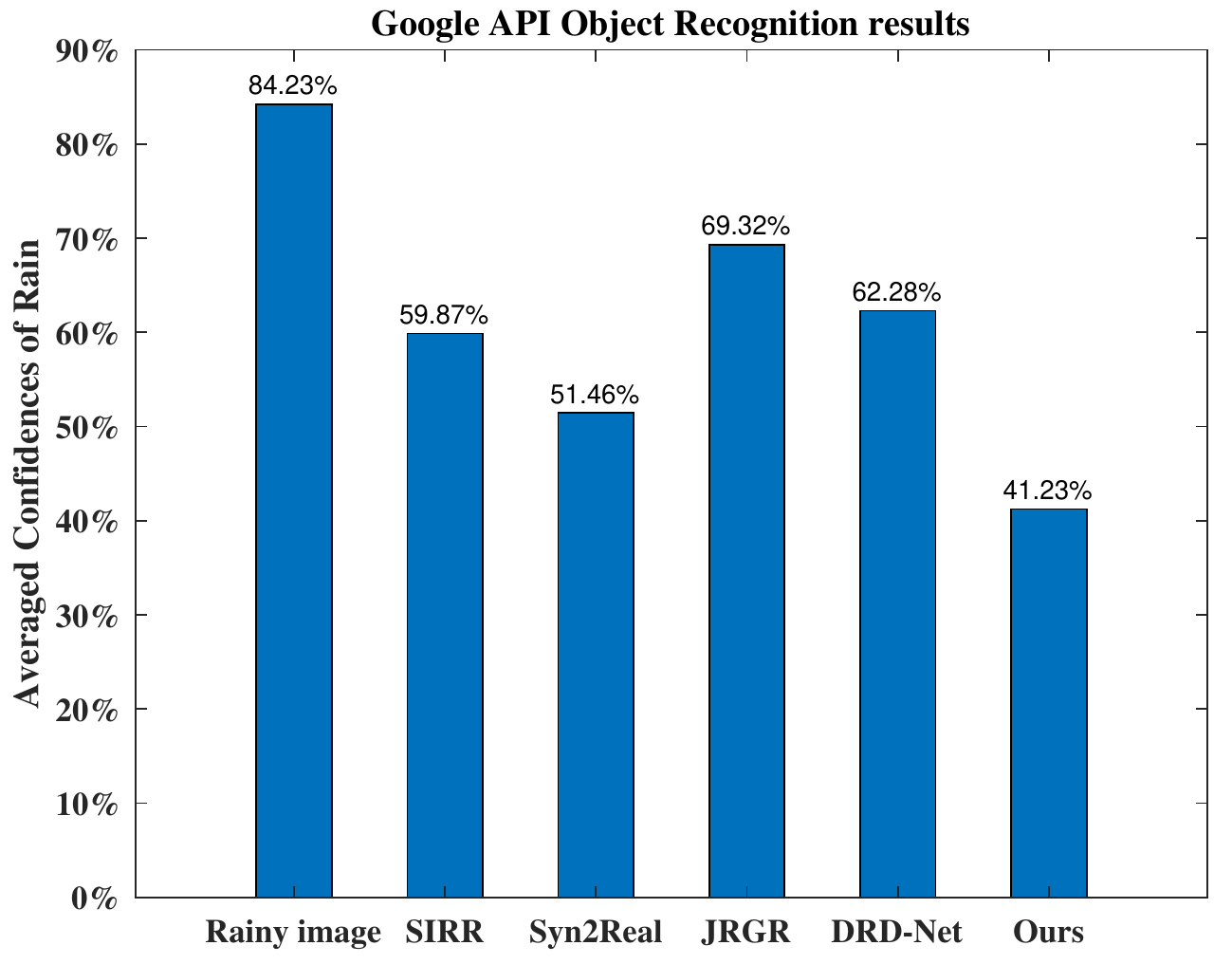}
	}
	\caption{Deraining results tested on Google Vision API. From (a)-(c): (a) the object recognition result in the real-world rainy image, (b) the object recognition result after deraining by our Semi-DRDNet, and (c) the averaged confidences in recognizing rain from 100 sets of the real-world rainy images and derained images of SIRR \cite{wei2019semi}, Syn2Real \cite{yasarla2020syn2real}, JRGR \cite{ye2021closing}, DRD-Net \cite{deng2020detail} and our Semi-DRDNet respectively. Note: zero confidence refers to a total failure in recognizing rain from a derained image by Google Vision API.}
	\label{fig:google_api}
\end{figure*}

\begin{figure*}[!t]
\centering
\subfigure[Real-world rainy image]{
\includegraphics[width=0.32\linewidth]{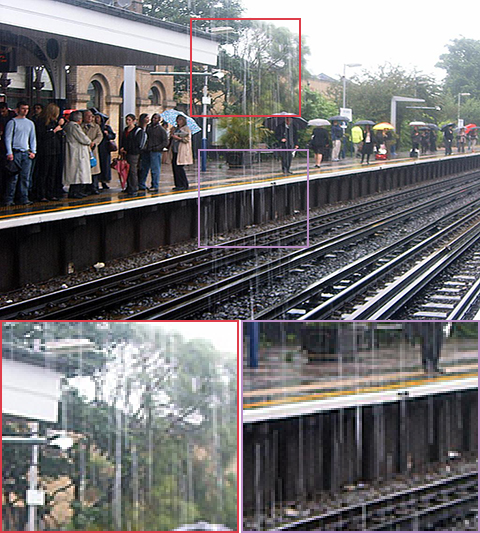}
}
\subfigure[MPRNet \cite{zamir2021multi}]{
\includegraphics[width=0.32\linewidth]{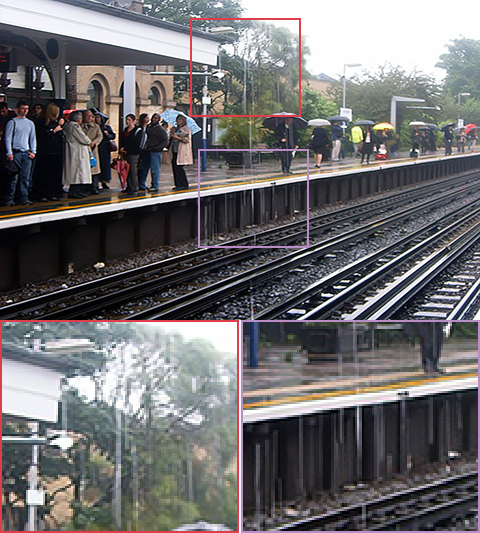}
}
\subfigure[MSPFN \cite{jiang2020multi}]{
\includegraphics[width=0.32\linewidth]{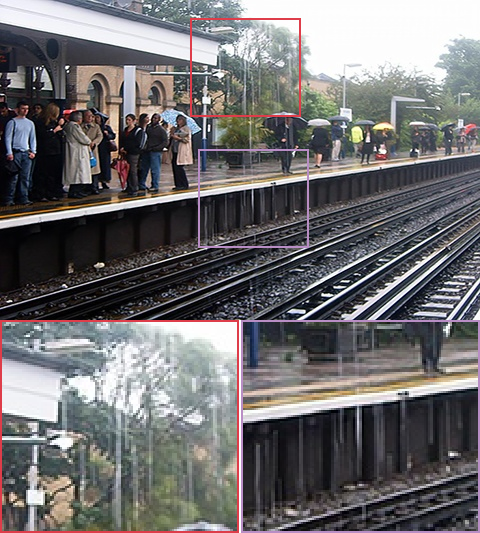}
}

\subfigure[Syn2Real  \cite{yasarla2020syn2real}]{
\includegraphics[width=0.32\linewidth]{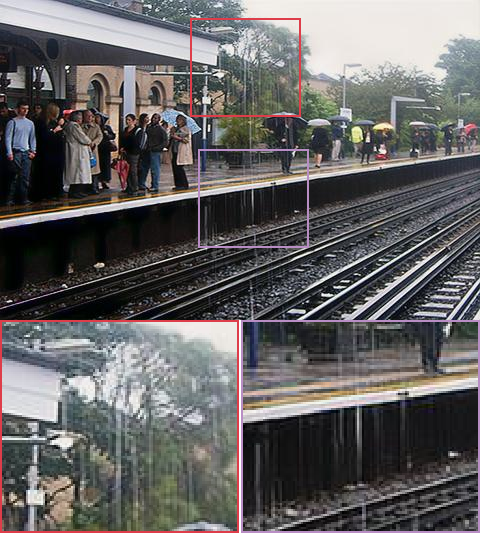}
}
\subfigure[DRD-Net \cite{deng2020detail}]{
\includegraphics[width=0.32\linewidth]{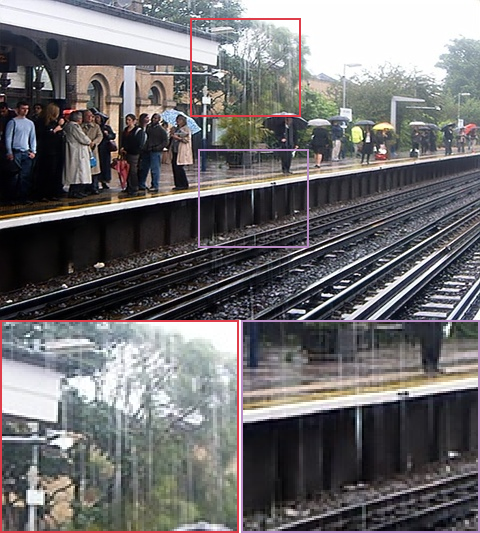}
}
\subfigure[Semi-DRDNet (ours)]{
\includegraphics[width=0.32\linewidth]{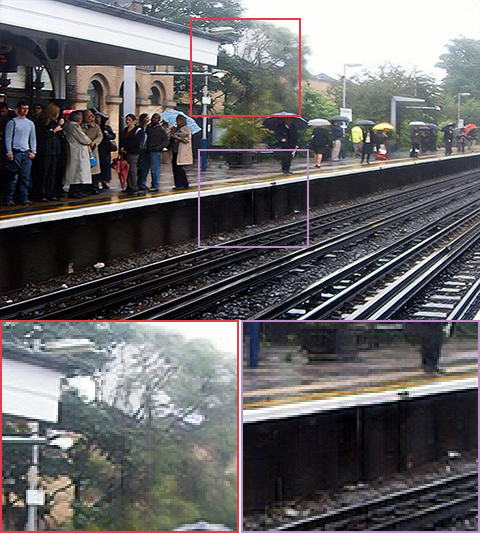}
}
\caption{Real-world image deraining in our established real-world dataset, called Real200. Although challenging to cope with real-world rainy images, Semi-DRDNet not only safeguards the result from rain remnants, but also provides the rain-affected areas with abundant real details, which is largely different from its (semi-)supervised competitors including our conference version, i.e., DRD-Net \cite{deng2020detail}.}
\label{fig:castle}
\end{figure*}


\subsection{Application} 
To demonstrate that our Semi-DRDNet can benefit vision-based applications, we employ Google Vision API to evaluate the deraining results. One of the results is shown in Fig. \ref{fig:google_api} (a-b) where Google Vision API can recognize the rainy weather in the rainy image while it cannot recognize the rainy weather in the derained image. Furthermore, we use Google Vision API to test 100 sets of the real-world rainy images and derained images of our Semi-DRDNet, DRD-Net \cite{deng2020detail} and three semi-supervised methods \cite{wei2019semi,yasarla2020syn2real,ye2021closing} in Fig. \ref{fig:google_api} (c). After deraining, the confidences in recognizing rain from the images are significantly reduced.

\subsection{Running Time}
We compare the running time of our method with different methods on Rain200H in Fig. \ref{Tab:time}. It is observed that our method is not the fastest one, but its performance is still acceptable.

\vfill

\end{document}